\documentclass[journal]{IEEEtran}
\usepackage{graphicx}
\usepackage{url}
\usepackage{algorithm}
\usepackage{float}
\usepackage{booktabs}
\usepackage{cite}
\usepackage{algpseudocode}
\usepackage{amsmath}
\usepackage[dvipsnames,svgnames,x11names,table]{xcolor}
\usepackage{pifont}
\usepackage{hyperref}

\usepackage{dblfloatfix}

\begin{document}

\title{Towards a Transparent and Interpretable AI Model for Medical Image Classifications}

\author{Binbin Wen, Yihang Wu, Tareef Daqqaq, Ahmad~Chaddad*
\IEEEcompsocitemizethanks{\IEEEcompsocthanksitem
B. Wen, Y. Wu and A. Chaddad are with the School of Artificial Intelligence, Guilin University of Electronic Technology, Guilin, China. \\*Corresponding author: Ahmad Chaddad. \\
A. Chaddad is with The Laboratory for Imagery, Vision and Artificial Intelligence, Ecole de Technologie Superieure, Montreal, Canada.\\
T. Daqqaq is with the College of Medicine, Taibah University, Madinah  
42361, Saudi Arabia.\\
T. Daqqaq is with the Prince Mohammed Bin Abdulaziz Hospital, Ministry of 
National Guard Health Affairs, Al Madinah, Kingdom of 
Saudi Arabia\\
Email: ahmad8chaddad@gmail.com, ahmadchaddad@guet.edu.cn}}

\markboth{Journal of \LaTeX\ Class Files,~Vol.~14, No.~8, August~2015}
{Shell \MakeLowercase{\textit{et al.}}: Bare Demo of IEEEtran.cls for IEEE Journals}

\maketitle

\begin{abstract}
The integration of artificial intelligence (AI) into medicine is remarkable, offering advanced diagnostic and therapeutic possibilities. However, the inherent opacity of complex AI models presents significant challenges to their clinical practicality. This paper focuses primarily on investigating the application of explainable artificial intelligence (XAI) methods, with the aim of making AI decisions transparent and interpretable. Our research focuses on implementing simulations using various medical datasets to elucidate the internal workings of the XAI model. These dataset-driven simulations demonstrate how XAI effectively interprets AI predictions, thus improving the decision-making process for healthcare professionals. In addition to a survey of the main XAI methods and simulations, ongoing challenges in the XAI field are discussed. The study highlights the need for the continuous development and exploration of XAI, particularly from the perspective of diverse medical datasets, to promote its adoption and effectiveness in the healthcare domain. Our code is available at \url{https://github.com/AIPMLab/XAI_-review-2024}

\end{abstract}


\IEEEpeerreviewmaketitle

\section{Introduction}\label{sec1}

With the rapid development of artificial intelligence (AI), especially deep learning (DL) and machine learning (ML) techniques, new opportunities for economic and social growth are emerging \cite{chaddad2025eamapg,hassan2023explaining,sivamohan2023optimized}. For example, the advent of ResNet \cite{he2016deep} increases the classification accuracy of deep models by large margins in ImageNet compared to AlexNet and the VGG series \cite{krizhevsky2012imagenet,simonyan2014very}. Furthermore, the use of transformers in vision (ViT) opens avenues for introducing techniques derived from different domains (e.g., natural language processing (NLP)) to improve classifier performance \cite{han2022survey}. Similarly, multi-modal pretraining can considerably improve zero-shot classification accuracy (e.g., contrastive language pre-training (CLIP) can achieve similar accuracy using zero-shot compared to a fully-trained ResNet50 on ImageNet2012) \cite{radford2021learning}. However, despite the remarkable performance of these deep models, they tend to be classified as black-box models, meaning that it is difficult to interpret the model decision-making process \cite{arrieta2020explainable,dwivedi2023explainable}. This presents a significant challenge, particularly in the healthcare field, leading to a trust gap between physicians and their patients \cite{dhar2023challenges}. In addition, radiologists may validate interpretations driven by \textcolor{black}{explainable artificial intelligence (XAI)}, ensuring that AI predictions align with clinical expectations and improve diagnostic accuracy.


Specifically, when the application of DL involves high-risk applications, such as cancer diagnosis, which are sensitive to technique and closely related to human survival or safety, its black-box nature can make it difficult for people to see the reasons for decision making \cite{hassija2024interpreting,el2023utilizing,jiang2025knowledge}.
If the predictions are not accurate, it will cause problems for human life \cite{dwivedi2023explainable}. In the medical domain, a scenario can be considered in which a DL model infers the presence of pulmonary tuberculosis in a patient based on test data. However, doctors must determine where the diagnosis results come from, so they dare not use them directly. Instead, they carefully checked the relevant test data according to their own experience and then made a judgment. To address these challenges, it is essential to understand the workings of black-box models. This involves making the models more transparent and clarifying the underlying principles. This highlights the need to enhance the transparency of black-box models to increase their reliability, which is the aim of XAI \cite{chaddad2023survey,chaddad2023enhancing}. 


Current academic research has made significant advances in the field of XAI. Techniques such as rule-based learning, visualization of learning processes, knowledge-based data representation, and human-centered AI models are being employed to improve the interpretability of AI systems \cite{hassija2024interpreting}. The applications of XAI are expanding across various fields, including smart cities \cite{sindiramutty2024modern}, intrusion detection \cite{sharma2024explainable,arreche2024xai}, and decarbonization technologies \cite{patwary2024explainable}, with healthcare being a prominent area of focus. Among these topics, the application and research of XAI in disease prediction, diagnosis, and treatment are the most extensive \cite{wen2023use}. For example, in \cite{ghnemat2023explainable}, a novel XAI method was introduced to segment images and provide a better understanding of how the AI model arrives at its results for chest x-ray image classification tasks. Similarly, in \cite{farrag2023explainable}, the gradient-weighted class activation map (Grad-CAM) was used to visualize the decision process of Deeplabv3 and improve the interpretability of breast cancer segmentation tasks.

\begin{figure}[ht] \centering
   
    \includegraphics[width=0.98\linewidth]{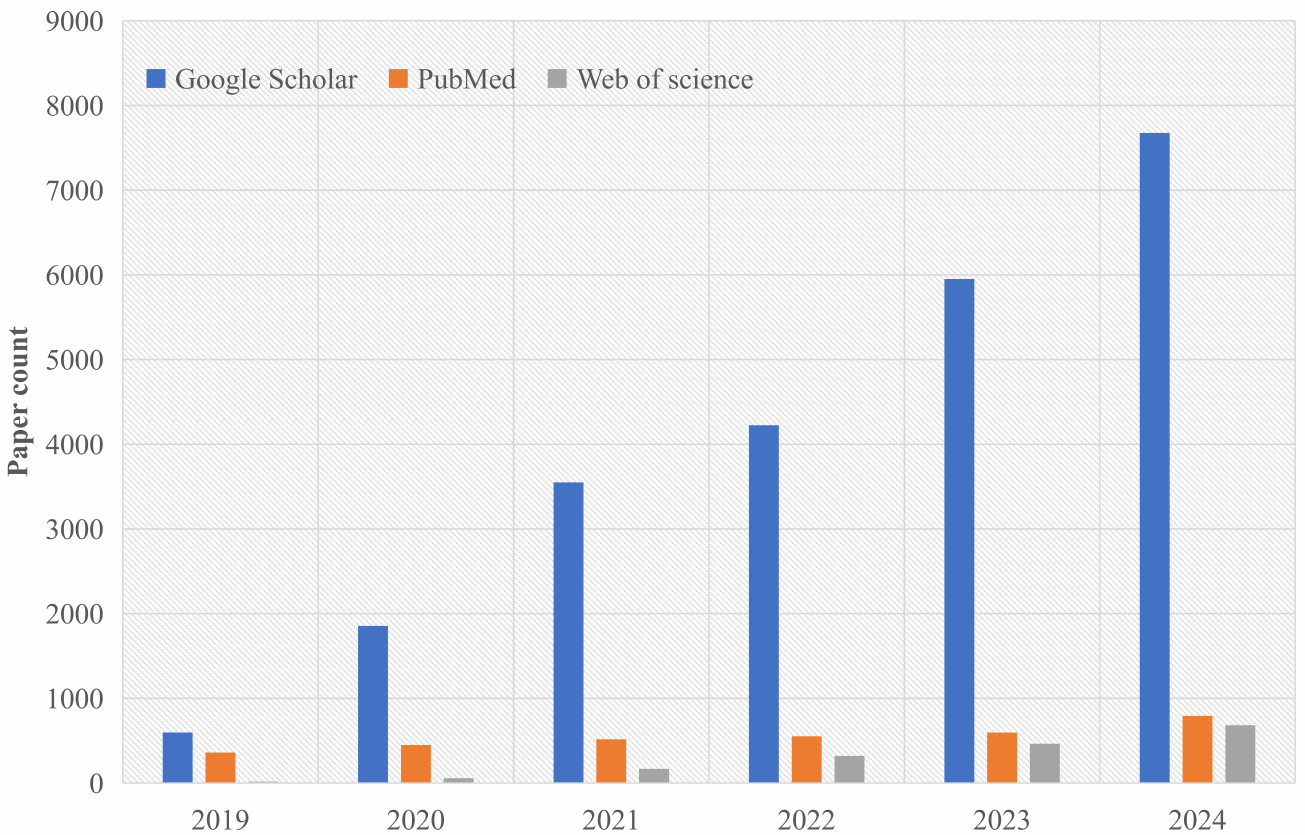}
    \caption{\textcolor{black}{Number of publications in the XAI domain over the last five years. The data was obtained from three distinct sources, namely Google Scholar, PubMed, and Web of Science. Data collection ends on 2025/3/1.}}
    \label{fig:my_label_14}
\end{figure}

To demonstrate the impact of XAI in research, we illustrate Figure \ref{fig:my_label_14} to show the number of papers related to “explainable AI in medicine” from 2020 to 2025, indexed in Google Scholar, PubMed, and Web of Science. As shown, XAI has attracted considerable attention from the academic community (e.g., approximately 6000 papers published in 2023). Compared to recent state-of-the-art (SOTA) surveys \cite{van2022explainable,chaddad2024generalizable,hossain2023explainable}, the novelty of this study is that we use common XAI techniques to explain model predictions on five publicly available medical datasets and introduce a quantitative metric to quantify the XAI approaches. Furthermore, we measured the execution time to provide a comprehensive analysis of XAI models.

The contributions of this paper are as follows.
\begin{itemize}
\item We provide a timely survey of XAI approaches especially in the medical image analysis domain, covering the common XAI-based classification methods, with the algorithms in their respective categories.
\item We perform simulation studies to show the decision progress of deep models using public medical datasets. Furthermore, we introduce two quantitative metrics (fidelity score and execution time) to evaluate the effectiveness of those XAI approaches. 
\item We discuss the potential challenges facing XAI in classifier models and future trends. 
\end{itemize}

The rest of this paper is structured as follows. In Section \ref{1}, we focus specifically on recent research publications in the last two years to provide a comprehensive overview of the most current advances in the field of XAI. Section \ref{2} discusses three widely used classification techniques in XAI. Section \ref{3} presents the concept of human-centered XAI, which includes classification for various demographics and the integration of human-centered design principles into XAI. In Section \ref{4}, we explore the use of XAI in healthcare, covering objectives targeted in this area, the integration of XAI with computer-aided diagnosis and electronic health record (EHR) systems, as well as the categorization of XAI solutions for healthcare assignments. Section \ref{5} presents a set of experiments conducted to demonstrate the effectiveness of XAI techniques in interpreting images classified by AI models. Section \ref{6} outlines the current challenges faced by XAI. Finally, Section \ref{7} provides the concluding remarks of this paper.

\begin{figure*}[htp] \centering
    \includegraphics[width=0.9\linewidth]{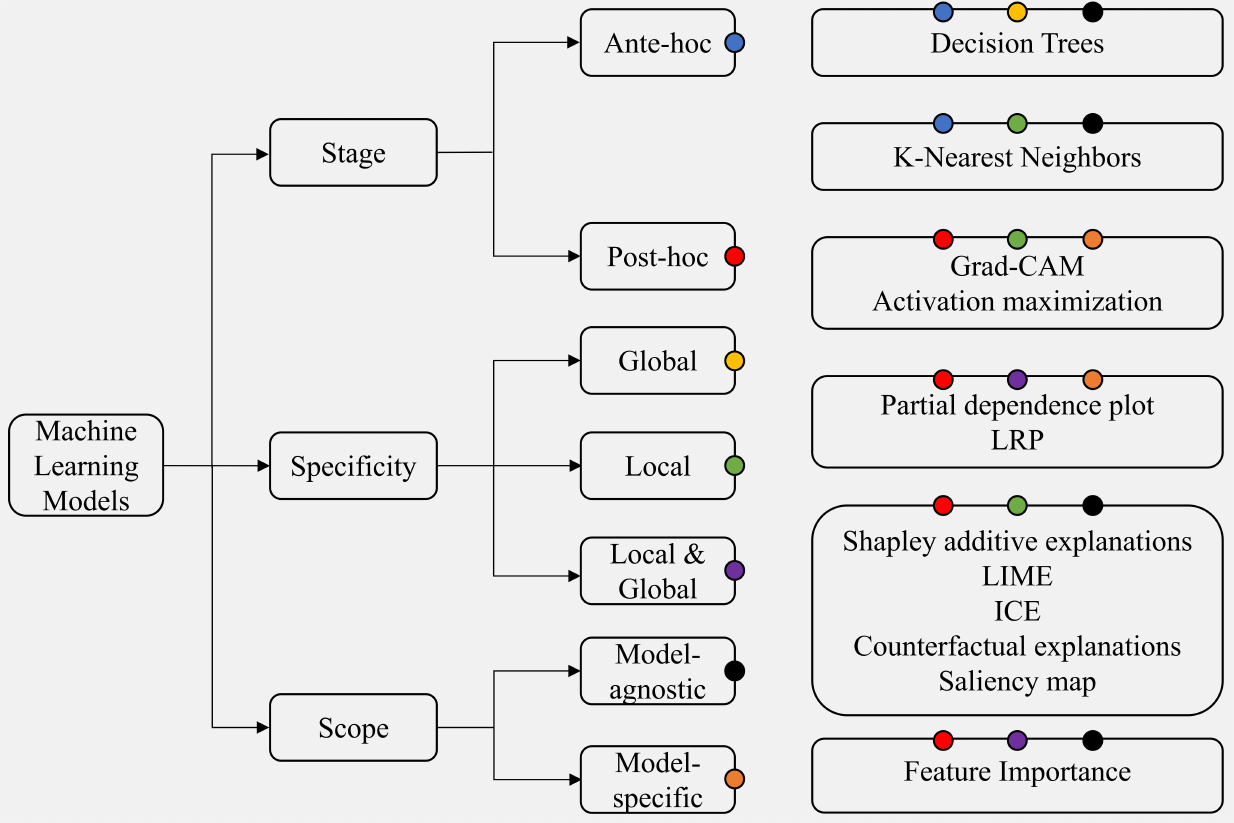}
    \vspace{10pt}
    \caption{Classification of XAI methods \cite{32}. {Classify the twelve common XAI methods based on the previous three methods (stage, specificity, and scope).} As shown in the figure, Post-hoc and model-agnostic contain more techniques than Ante-Hoc and model-specific, respectively. This can also indirectly reflect that Post-hoc and model-agnostic are more popular among users.}
    \label{fig:my_label 2}
\end{figure*}

\section{Related work} \label{1}
XAI provides transparency and interpretability for model decisions in medical applications. This enhances the credibility of the model and thereby supports clinical decision-making. For example, in \cite{samek2017explainable}, sensitivity analysis and layer-wise relevance propagation (LRP) were introduced to highlight progress in model decisions for medical image classification tasks. In \cite{van2022explainable}, an XAI framework was proposed based on global and local explanations and model-specific and model-agnostic approaches, offering a baseline approach to systematically interpret various techniques and demonstrate their potential in CT and MRI image analysis. Furthermore, in \cite{singh2020explainable}, Grad-CAM and attention techniques were introduced to improve the visual interpretability of deep models, highlighting advances in disease detection and diagnosis. These studies demonstrate the potential of XAI in improving model transparency, increasing trust, optimizing performance, and driving innovation in medical image analysis.

To improve the interpretability of the model, for example, in \cite{el2024exhyptnet}, they proposed to use the features extracted by the DL models with Grad-CAM interpreters to automatically detect hypertension by analyzing photoplethysmogram (PPG) signals. Experimental results in two public medical data sets demonstrate the effectiveness of their method. Furthermore, to diagnose Type II diabetes, a new XAI technique is introduced using explanations based on random forest (RF) rules \cite{chen2024deep}. The simulation results show a higher performance metric of the proposed model. In \cite{lalithadevi2023feasibility}, they suggested an XAI method based on RF combined with the Shapley additive explanations (SHAP) method to predict the risk of diabetic retinopathy (DR) in patients with type II diabetes. Moreover, in \cite{dharmarathne2024novel}, they developed a self-explanatory interface for the diagnosis of diabetes to provide a transparent explanation. Their approach highlights the value of early diagnosis and preventive care in the medical field, and their method can achieve high accuracy rates ranging from 73.52\% to 74.15\%.

In \cite{wani2024deepxplainer}, they proposed a new XAI method for the classification of lung cancer. The main idea is to add the XGBoost classifier to CNNs to provide precise predictions. The results of SHAP also indicate the effectiveness of their method. Furthermore, XAI methods such as SHAP, local interpretable model agnostic explanations (LIME), are used for the diagnosis of COVID-19, thus improving the transparency of the models \cite{khanna2023machine}. The experimental results show that their method can help healthcare professionals optimize resource allocation and reduce mortality rates. For cardiovascular diseases, combined SHAP with DL models is used to improve the interpretability of predictions \cite{hossain2023cardiovascular}. Moreover, in \cite{mahim2024unlocking}, they proposed using the Vision Transformer (ViT) with SHAP and LIME to improve the explainability of the classifier model using the Alzheimer's disease data set. Their method shows higher performance compared to other classification techniques. 

Recently, Nafisah et al. \cite{nafisah2024tuberculosis} introduced an automated tuberculosis detection system based on DL, focusing on extracting regions of interest from chest X-ray films (CXR) and using advanced CNN models for analysis. Using a complex image segmentation network and multiple pre-trained CNN models, such as EfficientNetB3, the system can automatically distinguish tuberculosis and non-tuberculosis images with a high accuracy rate of 99.1\%. Furthermore, using XAI techniques such as Grad-CAM and t-Distributed Stochastic Neighbor Embedding (t-SNE), the transparency of the model and the confidence of physicians in the diagnostic process are improved, confirming that this method can serve as an effective and rapid diagnostic tool with the potential to significantly reduce deaths caused by diagnostic delays or errors.

These studies collectively demonstrate the evolving landscape of AI in healthcare, highlighting the shift toward methodologies that improve predictive metrics and explainability, thus ensuring that AI-driven decisions are accessible and transparent to healthcare professionals and patients.

\section{Classification of XAI methods} \label{2}
XAI methods can be classified using different criteria, such as scope, data type, and stage, including Ante-Hoc and Post-Hoc classifications. This study focuses primarily on stage-, scope-, and specificity-based classification methods, as illustrated in Figure \ref{fig:my_label 2}. For example, the attributes of the decision tree are Ante-Hoc, both global and local, and model specific. Next, we provide a summary of the three different classification methods, the approaches shown in the diagram, and illustrative case studies.

\textbf{Ante-Hoc versus (vs.) Post-Hoc}: Ante-Hoc explainability refers to the ability to make the model itself explainable by training a model with simple structure and good explainability, or by integrating explainability into a specific model structure.

A self-explanatory model means that a model is considered transparent if it is itself understandable. Therefore, self-explanatory models are also called transparent models. Standard models are linear/logistic regression, decision tree, K-Nearest Neighbors (KNN), Bayesian Models, Rule-based, etc.

Post-Hoc explainability (also known as post-modeling explainability) refers to interpreting trained ML models by developing explainability techniques. Given a trained learning model, the goal of post-hoc explainability is to use explanation methods or build an explanation model to explain the decisions or behaviors made by the model.

\textbf{Global vs. Local}: Global explainability means understanding the overall decision logic of the model, while local explainability focuses on the interpretation of individual predictions.

Current XAI approaches, such as LIME and SHAP, focus on local interpretability, helping to understand how a specific prediction is made. This is very important in healthcare, where understanding individual predictions for each patient is important. However, global explanations look at the overall patterns of a model but might miss important details needed for personalized decision-making in complex and dynamic situations \cite{7}.

Suppose a DL model diagnoses a chest X-ray image as "pneumonia." Global explainability may disclose the reasons for predicting this class based on age, medical history, etc., but it cannot identify which areas in this specific X-ray image contributed to the diagnosis of "pneumonia." Local explainability directly analyzes specific areas in the image and observes which areas have the highest impact on the prediction results through local perturbations. It thereby identifies specific imaging features that may be related to pneumonia (e.g., pulmonary ground-glass shadow) \cite{fontes2024application}.

\textbf{Model-agnostic vs. Model-specific}: Whether a model is agnostic means that the model can be applied to many algorithms or can only be used for a specific model. Model-specific interpretation tools are intrinsic model-interpretation methodologies based entirely on each model. The model-agnostic approach has no requirements for a model or algorithm. It works with any ML model and is often used with Post-Hoc explanations. Although model-specific, this approximate one-to-one explanation requires high predictiveness, and combined with its specificity determines its narrower application. As a result, researchers have recently become more interested in model-agnostic approaches.

\textbf{Decision trees (DT)}: It is an instance-based inductive learning algorithm that divides unordered samples into different branches according to certain rules based on the characteristics of the samples to achieve classification or regression purposes. Decision trees can be used to provide useful visual explanations of decision results and are considered the simplest and most interpretable white-box ML algorithm in human decision-making processes \cite{31}. Furthermore, DT algorithms are easy to understand, and the complexity of DT is only influenced by the number of layers in the decision tree. Moreover, the data processing efficiency is high, which makes them suitable for real-time classification scenarios.

\textbf{Grad-CAM}: Grad-CAM is a gradient-based visualization method \cite{selvaraju2017grad}. It is a generalized form of CAM, which can be combined with various CNN models. As long as the model has convolutional layers, Grad-CAM can be used to visualize its decision-making process. The advantage of Grad-CAM is that the generated map preserves the spatial information of the input image, allowing precise positioning of specific areas in the image related to the predicted results \cite{10085971}. Grad-CAM visualization can be defined as follows.
\begin{equation}
    \omega_k = \frac{\sum_{i,j} \frac{\partial y}{\partial A_{ij}}}{\sum_{i,j} \left(\frac{\partial y}{\partial A_{ij}}\right)^2 + \left(A_{ij}\right)^2 \cdot \sum_{i,j} \left(\frac{\partial y}{\partial A_{ij}}\right)^2} + \epsilon
\end{equation}
where $A_{ij}$ are elements of the feature map, $y$ is the class score, and $\epsilon$ is a small constant to prevent division by zero.
\begin{equation}
    \text{Grad-CAM} = \text{ReLU}\left(\sum_{k} \omega_k \cdot A_k\right)
\end{equation}
where $\omega_k$ are the weights for the $k$-th channel, and $A_k$ is the feature map for the $k$-th channel.

\textbf{Activation maximization (AM)}: Its idea is to find the input mode that maximizes the activation value of a given hidden layer unit and to find an input that needs to maximize the output value of the neuron to be explained \cite{katzmann2021explaining}. However, the AM method is more suitable for optimizing continuous data than for discrete data such as text, graph, data, etc., so it is difficult to directly interpret NLP models and graph neural network (GNN) models. However, the AM method is less capable of processing discrete data (e.g., text, graph, etc.), and it is not widely used for NLP and GNN tasks. The core of AM is to find the following activation:
\begin{equation}
    \max_{x} \, A^l(x; \theta)
\end{equation}
where $A^l(x;\theta)$ is the activation of $l$-th layer given input $x$ and model parameters $\theta$ (e.g., weights).


\textbf{K-nearest neighbors (KNN)}: While the concept of KNN is straightforward, a significant challenge is its inability to provide a deep understanding of the data. However, the interpretability of KNN is inherently robust since the closest neighbors offer explanations that are easily comprehensible to humans \cite{31}. Therefore, the transparency of KNN depends in part on the number of neighbors \cite{17}.

\textbf{SHAP}: The goal of SHAP is to explain the prediction of an instance $x$ by calculating the contribution of each feature to the prediction $x$ \cite{antwarg2021explaining}. SHAP can be seen as an interpretive prediction framework that unifies these local interpretive methods (LIME), DL important features (DeepLIFT), layer-wise release promotion, Shapley compression values, Shapley sampling values and quantitative input influence), allowing them to be applied more consistently to different ML models and tasks, thus improving the consistency and comparability of interpretive performance \cite{lundberg2017unified}. Typically, the SHAP value $\phi_i(v)$ can be calculated using the following equations:
\begin{equation}
    \phi_i(v) = \sum_{S \subseteq N \setminus \{i\}} \frac{|S|!(n-1-|S|)!}{n!} \left( f(v_{S \cup \{i\}}) - f(v_S) \right)
\end{equation}
where $N$ is the set of all features, $S$ is a subset of $N$ without feature $i$, $v$ is the set of feature values, $f$ is the model prediction function, and $n$ is the total number of features. $\phi_i(v)$ indicates the specific SHAP value for the feature $i$.


\textbf{LIME}: The idea is to explain the prediction of complex models (such as deep neural networks \textcolor{black}{(DNNs)}) by fitting a local surrogate model. The prediction of this model is easy to explain \cite{84}. One of the advantages of this method is its ability to provide good performance in tasks related to image classification and NLP. Because LIME itself is model independent, it has a wide range of applicability. However, due to poor stability and repeated operations, the samples generated by the disturbance may differ widely, and the explanations given may vary greatly. The core idea of LIME can be represented as follows.
\begin{equation}
    \text{LIME} = \arg\min_{g \in \mathcal{F}, \, g(x_0) = y_0} \sum_{x \in \Pi} \alpha \cdot d(x, x_0) + \lambda \cdot \mathcal{L}(f, g, x)
\end{equation}
where $g$ is the explanation model, $\mathcal{F}$ is the function class of the explanation model, $y_0$ is the prediction of the original model $f$ at $x_0$, $\Pi$ is a set of perturbed samples around $x_0$. $\alpha$ and $\lambda$ are the regularization terms. $\mathcal{L}$ is a loss function that measures the consistency between $f$ and $g$.

\textbf{Saliency map (SM)}: The saliency map is an image that highlights the parts of an image that are easily noticeable to the human eye, and its pixel values reflect the degree of most attractive visual attention to the human eye \cite{brahimi2018deep}. It represents the visual saliency of the corresponding visual scene and belongs to the attribution method, which can further improve the quality of visualization. Saliency maps are suitable for tasks such as determining visually significant areas, improving the accuracy of object detection, and image segmentation. Basically, the SM can be defined as:
\begin{equation}
    \text{Saliency Map} = \left| \frac{\partial y}{\partial x} \right|
\end{equation}
where $y$ is the output of the model (e.g., probability), $x$ is the input (e.g., image), $\frac{\partial y}{\partial x}$ represents the gradient.

\textbf{Individual conditional expectation (ICE)}: The ICE plot distinguishes from the partial dependency plot (PDP) by depicting the relationship between the predicted values of each individual and a single variable \cite{goldstein2015peeking}. However, it needs to follow the assumption of independent variables and thereby decreases its potential for some tasks that cannot meet the requirements. Furthermore, it cannot reflect the relationship between multiple feature variables and the target. Unlike PDP, which shows the average effect of a set of features, the ICE plot eliminates the influence of non-uniform characteristics and visualizes the dependence of the prediction of each sample on the features separately. One advantage is that it can help to solve the problem of data heterogeneity and is easy to put into practice, but due to the large number of individuals in the ICE plot, it can make the image cumbersome and difficult to obtain an explanation. The ICE process can be defined as:
\begin{equation}
\begin{split}
\text{For } i \in \{1, 2, \ldots, N\}, \text{ plot the curve } \hat{f}_S^{(i)} \text{ against } x_S^{(i)}, \\
\text{ while } x_C^{(i)} \text{ remains fixed.}
\end{split}
\end{equation}
where $i$ represents each instance, $\hat{f}_S^{(i)}$ indicates the predicted output.

\textbf{Layer-wise relevance propagation (LRP)}: LRP is a propagation-based interpretation method that requires access to the interior of the model, such as topology, weights, and activation \cite{montavon2019layer}. LRP uses the network structure and redistributes explanatory factors, starting from the output layer of the network and propagating back to the input layer. Each redistribution can be seen as a simple solution to explain the problem. The main advantage of LRP is its high efficiency in computation, while its disadvantage is poor flexibility. Specifically, let $j$ and $i$ be neurons at two consecutive layers of the neural
network, the LRP can be represented as:
\begin{equation}
    R_j = \frac{\sum_{i} z_{ij} \cdot R_i}{\sum_{i} |z_{ij}|}
\end{equation}
where the relevance $R_j$ is determined by the weighted sum of the relevance $R_i$, $z_{ij}$ represents the contribution of the i-th neuron in the preceding layer to the j-th neuron in the current layer.

These XAI techniques are summarized in Table \ref{T:4}. Specifically, we report the advantages, disadvantages, and suitable applications to guide the selection of the appropriate technique for specific tasks.

\begin{figure}[htp] \centering
   
    \includegraphics[width=0.98\linewidth]{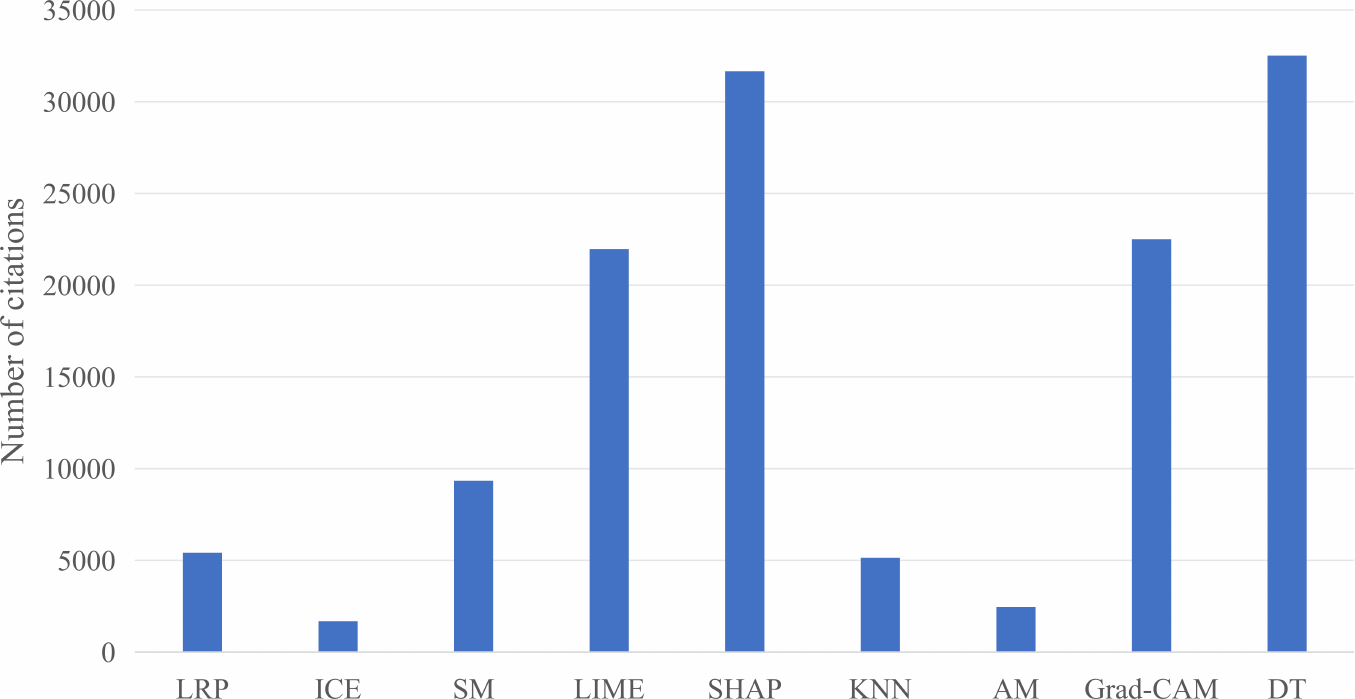}
    \caption{\textcolor{black}{Bars represent the citations of XAI based technique. We searched Google Scholar for citations of the original papers for the nine techniques mentioned earlier. Data collection ends on 2025/3/1.}}
    \label{fig:my_label 3}
\end{figure}

Also we have collected as many references as possible for the techniques mentioned above, as shown in Figure \ref{fig:my_label 3}. We can see that, in addition to the basic decision tree technology, the most popular techniques are SHAP, LIME, and Grad-CAM, and this conclusion can also be proven in the papers we have collected.

\begin{table*}\scriptsize
    \caption{Comparative summary of XAI techniques for interpretability and application scenarios.}
    \begin{tabular}{m{2cm} | m{1.1cm}| m{1.4cm}| m{1.4cm}| m{1.4cm} |m{1.2cm} |m{1.1cm}|m{1.2cm} |m{3.2cm}}
    \hline
      \textbf{Methods} & \multicolumn{4}{c|}{\textbf{Advantages}} & \multicolumn{3}{c|}{\textbf{Disadvantages}} & \textbf{Suitable applications} \\ \hline
      
      & Easy to interpret & Support for unstructured data & Highly \ \  reproducible results & Low data preprocessing requirements & High computational complexity & Sensitive to data & Low robustness to noise \\ \hline
      
      DT \cite{quinlan1986induction} & \ding{51} &  & \ding{51} &\ding{51} &  &  & & Small datasets; rule-based problems (e.g., medical, business) \\ \hline
      
      Grad-CAM \cite{selvaraju2017grad} &  & \ding{51} &  &  & \ding{51} &  & & Visualizing Neural Network Models \\ \hline

      AM \cite{yosinski2015understanding} &  & \ding{51} &  &  & \ding{51} &  & \ding{51} & Feature visualization and  neuronal activation analysis \\ \hline
      
      KNN \cite{fix1985discriminatory} & \ding{51} &  &  & \ding{51} &  &  & & Smaller feature spaces and simpler data distributions \\ \hline

      SHAP \cite{lundberg2017unified} &  & \ding{51} & \ding{51} &  &\ding{51}&\ding{51}& \ding{51} & Tasks requiring feature contribution explanation \\ \hline
     
      LIME \cite{ribeiro2016should} & \ding{51} & \ding{51} & \ding{51} &  & \ding{51} & \ding{51} &\ding{51} & Explaining predictions of complex black-box models\\ \hline

      SM \cite{simonyan2013deep} &  &  &  &  &  & \ding{51} & \ding{51} &Require to know how the model interprets key areas of an image\\ \hline

      ICE \cite{goldstein2015peeking} &  &  &  &  & \ding{51} &  & & Best for understanding feature interactions, especially in regression models \\ \hline
      
      LRP \cite{bach2015pixel} &  & \ding{51} &  &  & \ding{51} &  & & Pixel-level explanations in CNNs \\ \hline
    
    \end{tabular}
    \label{T:4}

\end{table*}

The XAI methods presented in this section represent the most common tools used in the field. In medical images, the most appropriate XAI method is selected based on the nature of the data, the characteristics of the model and the depth of the explanation required. This evolving field will continue to support the transparency and reliability of ML models.



\section{Human-centered XAI}  \label{3}
Every product or tool is designed with a specific purpose or target audience in mind, and XAI is no exception. XAI has been developed to elucidate models that are beyond the scope of human comprehension.

\subsection{Classification and the reasons} 
\begin{figure}[htp] \centering
   \includegraphics[width=0.98\linewidth]{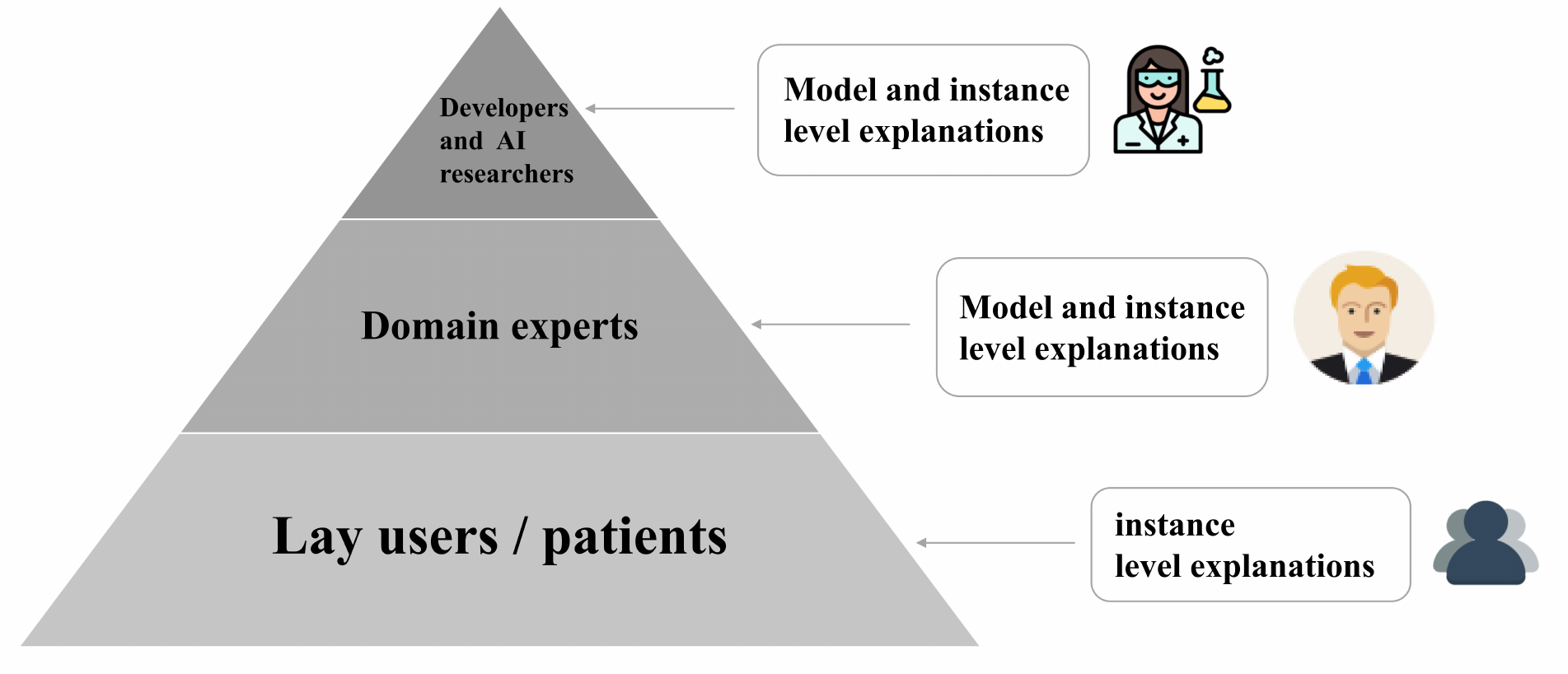}
    \caption{Categorize users based on their knowledge background and needs, and provide corresponding explanations. We have classified users into three groups and provided appropriate explanations tailored to their respective categories.}
    \label{fig:my_label 4}
\end{figure}

To better explain the different user groups, people/users are divided into different target groups and build particular explanations for them. Based on recent literature, we summarize these target groups as illustrated in Figure \ref{fig:my_label 4}.

Daudt et al. \cite{75} categorized users as experts or laymen, with the latter being end users who use AI products in their daily lives, but possess little or no knowledge of ML systems. On the other hand, Ribera et al. \cite{8} categorized people into developers and AI researchers, domain experts and lay users, with a focus on the medical field. Patients were not initially included in their target group, but with the emergence of data acquired using wearable devices, this may soon change. \textcolor{black}{Larasati et al. \cite{76} developed frameworks for explaining breast cancer diagnoses tailored to patients, non-experts, and AI users, ensuring comprehensibility based on each group knowledge and needs.}


The rationale for these varying explanations is as follows:
\begin{enumerate}
\item The degree of interpretation needed for a model is also dependent on the professional knowledge and capability of the user \cite{9}.
\item \textcolor{black}{key participants have varying objectives, requiring customized explanations.}
\item Certain explanations may be of greater significance to specific users.
\item The theoretical framework can guide the design of solutions to meet the specific needs and goals of users \cite{kim2024stakeholder}.
\item The identity of a person can change with changes in their environment.
\end {enumerate}


\subsection{Human-centered design for XAI} There are many human-centered designs, including media and entertainment, education, recommendation systems, healthcare, etc. We focus mainly on the healthcare field.

In an effort to improve user trust and comprehensibility, several studies have been proposed. Daudt et al. \cite{75} compared SHAP, LIME, and ranking importance to help understand the diagnostic results of cervical cancer. Lee and Rich \cite{77} evaluated the degree of trust a test group had in the decision to detect skin cancer made by AI and clinicians through online tests. \textcolor{black}{Wang et al. \cite{82} developed a theory-driven framework for an interpretable early decision support system to diagnose patients in intensive care units (ICUs). Furthermore, in \cite{81}, they collected insights from medical staff on prioritizing data during diagnosis and proposed essential conditions for medical assistant diagnosis systems.}

To improve transparency, \textcolor{black}{Cai et al. \cite{80} interviewed 21 pharmacologists before, during, and after the implementation of the prediction of prostate cancer diagnosis using DNN and learned the information and types of AI auxiliary system they wanted. Eiband et al. \cite{78} developed a step-by-step design method to build clear interfaces for different users.}

The human-centered approach to interpretation is more specific in terms of direction and purpose, which enables each target group to obtain more targeted and understandable explanations, resulting in greater satisfaction with the interpretation.

\section{Application of XAI in healthcare} \label{4} 
\textcolor{black}{XAI helps improve trustworthiness, interaction, and privacy awareness in healthcare by providing transparency and interpretability to AI systems.} For example, integration of AI and medical treatment has the potential to produce substantial benefits, such as cost reduction, increased efficiency, improved medical standards, and targeted decision-making support. The absence of transparency and interpretability in AI remains a significant challenge for its straightforward implementation in clinical settings. Therefore, the use of XAI related to medical will further promote the application of AI in this field.

\begin{table*}[htp] \scriptsize
    \setlength{\tabcolsep}{1.3mm}
    \caption{Summary of XAI Application in healthcare in 2023}
    \begin{tabular}{m{2.4cm}  m{2.3cm} m{2.4cm} m{1cm} m{6cm} m{2.5cm} } 
     \toprule
        \textbf{References} & \textbf{Diseases} & \textbf{Modality}&\textbf{XAI} & \textbf{Highlight/Contribution}&\textbf{ACC/PRE/SEN}\\
        \midrule
            Niranjan et al. \cite{88}& COVID-19 & CT & $\bullet$ & Guided Grad Cam-based Explainable Classification and Segmentation system &  98.51\ / 98.9(P), 98.0(N) / 98.2(P), 98.8(N) \\
           
           Yagin et al. \cite{86} & COVID-19 & Gene biomarkers & $\star$, $\diamond$ & Identify genetic biomarkers, sequencing (mNGS) samples & 93\ / - / 93.3\  \\
           
           Basahel et al. \cite{basahel2023quantum} & COVID-19 & X-ray & $\diamond$ &  Integration of quantum computing concepts in differential evolution & 97.43\ / 97.95\ / 96.88\ \\
           
           Seethi et al. \cite{seethi2024explainable}& COVID-19 & Gargle samples & $\diamond$, $\triangle$ & Using mass spectrometry data  & 94.12\ / 93\ / 93\ \\
           
           Chadaga et al. \cite{chadaga2023decision} & COVID-19 & Blood test & $\star$, $\diamond$ & Multi-level stacked model &  94\ / 95\ / 94\ \\
           
           Kirbouga et al. \cite{kirbouga2023cvd22} & COVID-19 & Blood test & $\diamond$ & Troponin levels & 83\ / 86\ / 80\ \\
           
           Elmannai et al. \cite{elmannai2023polycystic} & POS & BAD and CF & $\diamond$ &  Stacking ML with RFE & 100\ / 100\ / 100\ \\
           
           Khanna et al. \cite{khanna2023distinctive} & POS & BAD and CF & $\star$, $\diamond$ &  Multi-stack of ML models  & 98\ / 97\ / 98\ \\
           
           Hehr et al. \cite{87} & Leukemia & Blood smears & $\clubsuit$ & Single-cell based explainable multiple instance learning algorithm & - / - / - \\

           Aghaei et al. \cite{91} & Alzheimer's Disease & MRI & $\ast$ & Extracted 3D CNN based on ROI & 98.6\ / 98\ / 99.2\ \\
           
           Zhang et al. \cite{90} & Helicobacter Pylori & Endoscopy & $\ast$ & Feature extraction, combine LSTM and ResNet-50 &  91.9\ / - / 90.3\ \\
           
           Tasin et al. \cite{tasin2023diabetes} & Diabetes & BAD and CF & $\star$, $\diamond$ & Compare DT, SVM, RF, LR, KNN, and Ensemble Methods & 81\ / 81\ / 81\ \\
           
           Albahr et al. \cite{albahri2023explainable} & ASD & BAD and CF & $\star$ & Fuzzy approach-based multi-criteria decision-making &D1: 97.76\ / 97.76\ / 97.76\; D2: 78.12\ / 78.26\ / 78.12\ \\
           
           Das et al. \cite{das2023xai} & Heart diseases & BAD and CF  & $\star$, $\diamond$, $\spadesuit$ & Reducing dimensionality without reducing accuracy with XAI & 97.86\ / 96.95\ / 98.76\ \\
           
           Dong et al. \cite{dong2023explainable} & Gastric neoplasms & Endoscopy & $\ast$, $\star$ & Feature-extraction and multi-feature-fitting methods & 77.17\ / - / 87.3\ \\
           
           Auzine et al. \cite{auzine2023classification} & GC & pathological findings & $\diamond$ & Ensemble InceptionV3, Inception-ResNetV2 and VGG16 & 93.17\ / 97.2\ / 97.2\ \\
           
           Volkov et al. \cite{volkov2023possibilities} & Glaucoma & Fundus images & $\star$ & Inception V3, VGG 16, and ResNet 50 for binary classification of retinal images & 93\ / - / - \\
           
           Adilakshmi et al. \cite{adilakshmi2023medical}& ASD & BAD and CF & $\circ$ & System for diagnosing ASD using DT and SVM & 86\ / 88\ / 84\ \\
           
           Ahmed et al. \cite{ahmed2023identification} & Brain tumor & MRI &$\surd$ & Ensemble VGG-16 and LRP & 97.33\ / 99\ / 95.81\ \\

           Ramirez et al. \cite{ramirez2023explainable} & Prostate cancer & Gene expression & $\diamond$ & RF combined with class downsampling & - / - / 90\ \\

         \bottomrule
          \end{tabular}   
          \label{T:1}
          
Grad-CAM$\ast$: Gradient weighted class activation mapping; Guided Grad-CAM$\bullet$; LIME$\star$: Local Explainable Model Agnostic Explanation; SHAP $\diamond$: Shapley additive explanations; Permutation feature importance$\triangle$; Feature importance$\circ$; Self-explanation$\clubsuit$; PDP$\spadesuit$: Partial dependency plot; POS: Polycystic ovary syndrome; GC: Gastrointestinal Cancer; LRP$\surd$: Layer-wise relevance propagation; RFE: Recursive feature elimination; DT: Decision Tree; SVM: Support Vector Machine; RF: Random Forest; LR: Logistic Regression; KNN: K-Nearest Neighbors; ASD: Autism Spectrum Disorder; BAD: Biological assay data; CF: clinical features; P: Positive, N: Negative.
 \end{table*}


\subsection{Goals} \textcolor{black}{The integration of XAI techniques into the healthcare sector emphasizes specific objectives within the established XAI framework, including trustworthiness, interaction, and privacy awareness. These goals are crucial for ensuring that AI-driven decisions are accessible and transparent to healthcare professionals and patients.}

\subsection{XAI in computer aided diagnosis (CAD)} 
\begin{figure}[h] \centering
   \includegraphics[width=0.98\linewidth]{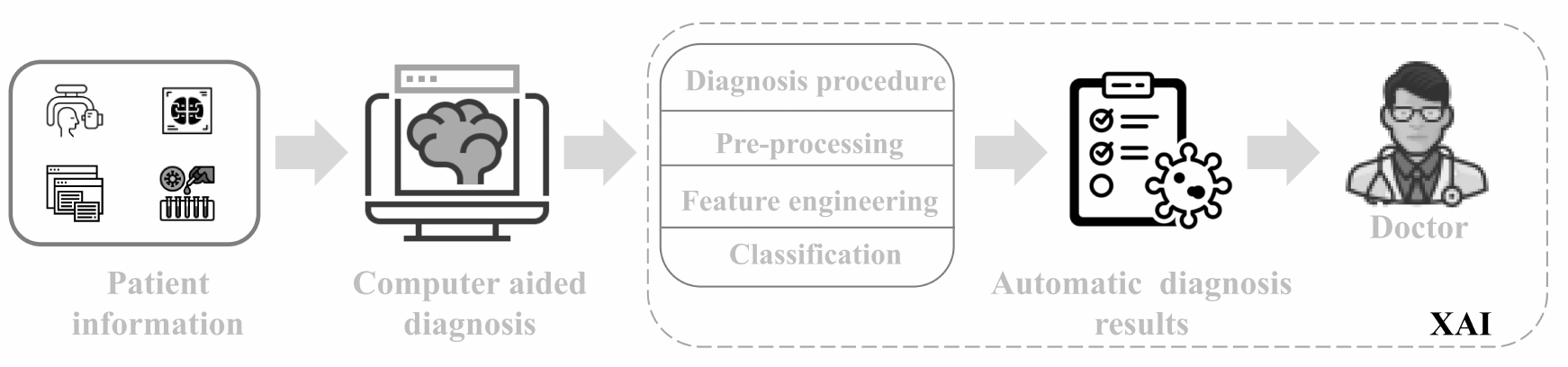}
    \caption{Flowchart of XAI assisting in CAD. It represents how we collect patient information through various methods and obtain the required features through feature engineering. These features are then sent to XAI, and with the help of XAI experts and clinicians, we ultimately obtain decision-making results such as prediction and diagnosis.}
    \label{fig:my_label 5}
\end{figure}

\textcolor{black}{AI technology has seen increasing application in various medical fields, including surgical robots \cite{panesar2019artificial}, medical image analysis \cite{england2019artificial}, medical decision-making \cite{reverberi2022experimental}, and personal medical assistants \cite{croatti2019bdi}. Efforts are being made to enhance the complexity and accuracy of CAD systems while prioritizing model interpretability in healthcare. This dual focus aims to improve clinical outcomes by supporting the development of appropriate treatment strategies, drug prescriptions, and dosage determinations.} Figure \ref{fig:my_label 5} shows the process of XAI assisting in CAD.

\textcolor{black}{A significant challenge in medical image analysis is the need for in-depth interpretive analysis to assist diagnosis and decision support.} In \cite{26}, they achieved explainable CAD by combining knowledge-driven and data-driven approaches. This integration can leverage the interpretability of the former and the high accuracy of the latter, leading to progress towards achieving multimodal diversity and complete interpretation that can be communicated to humans using understandable modalities such as text and concepts. According to \cite{27}, these advances are expected to significantly facilitate the adoption of XAI in assisted diagnosis applications.

\subsection{Electronic health records with XAI} 
\begin{figure}[htp] \centering
   \includegraphics[width=0.98\linewidth]{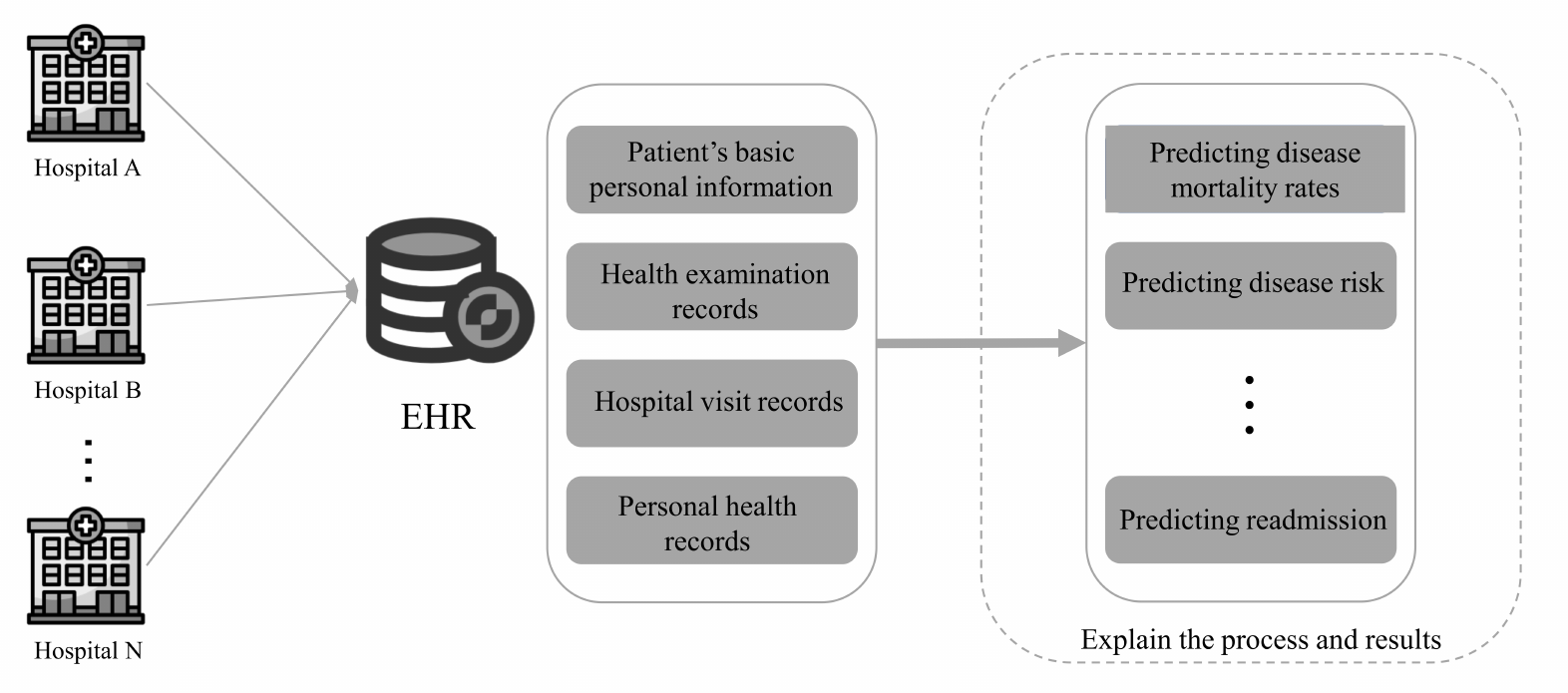}
    \caption{Workflow of EHR within XAI. EHR collects the stored information of patients in different hospitals to obtain patient records, and combined with XAI, provides explanations for disease prediction, readmission prediction, and other applications.}
    \label{fig:my_label 6}
\end{figure}

\textcolor{black}{EHR systems aim to improve the quality of medical services by enabling rapid access to a patient complete medical history, enhancing the accuracy and reliability of medical prescriptions, and facilitating efficient treatment \cite{hoerbst2010electronic}. A key feature of EHR systems is their interoperability, allowing healthcare professionals to generate, manage, and share digital health information across different healthcare entities, ensuring continuity and consistency in patient care.} Figure \ref{fig:my_label 6} illustrates a demonstration of the XAI workflow in the EHR context.


Currently, there exist several prediction models, including random forests, boosting trees, Naive Bayes, logistic regression, and decision trees. Despite their interpretability, these models rely on aggregation properties and do not account for the temporal correlation of the intrinsic properties of the EHR data, resulting in a suboptimal model accuracy \cite{khedkar2019explainable}.

A recent relevance study interprets the data in another way, comparing the interpretations given by XAI methods (SHAP, LIME, and Scoped Rules) as a third extension to analyze complex EHRs. Finally, the interpretation of the EHR through the XAI model can clarify the importance of features, providing important information for clinical decision-making \cite{duell2021comparison}.

In another study, Lauritsen et al. proposed an XAI early warning score (xAI-EWS) through hierarchical correlation propagation, which is used to predict severe acute disease from EHR. The system not only has high prediction performance, but can also provide clinicians with visual interpretation and prediction based on relevant EHR data so that clinicians can understand the potential causes of prediction \cite{lauritsen2020explainable}.

\subsection{Solution taxonomy} 
\begin{figure}[h] \centering
   
    \includegraphics[width=0.98\linewidth]{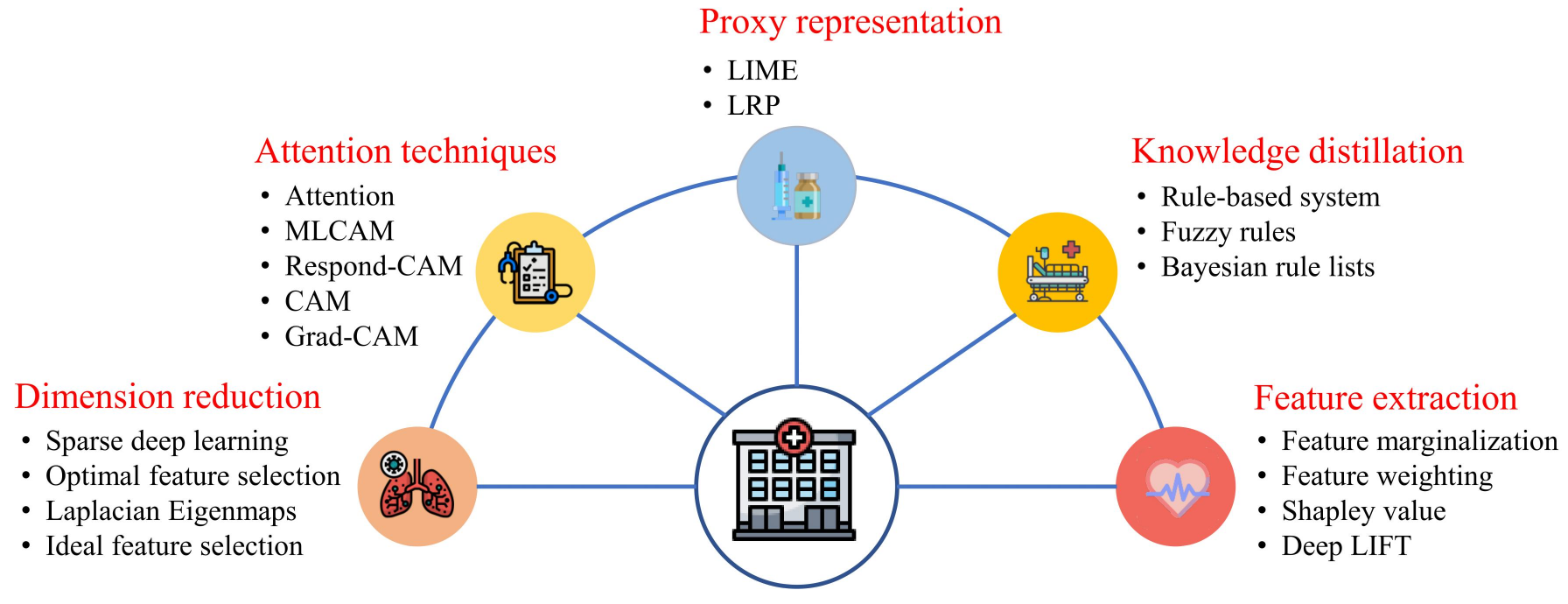}
    \caption{Different tasks in healthcare are classified according to the type of XAI interpretation technology \cite{31}. Under five different task classifications, we also provided several commonly used techniques for different tasks.}
    \label{fig:my_label 7}
\end{figure}

As shown in Figure \ref{fig:my_label 7}, this section discusses the solution taxonomy of XAI in healthcare by including different features for diagnosis and surgery.

\textbf{Dimension reduction}: \textcolor{black}{It aims to keep dominant features while eliminating redundant ones, reducing data storage space, computation time, and multicollinearity problem. For example, Elmannai et al. \cite{elmannai2023polycystic} developed an early detection model for polycystic ovary syndrome (PCOS) using ML and ensemble learning techniques. By combining different ML models, recursive feature elimination (RFE), and Bayesian optimization, they achieved high accuracy in early PCOS diagnosis. XAI enhances model transparency and usability in medical practice by providing insights at both model and individual levels.}


\textbf{Attention techniques}: \textcolor{black}{Similar to human visual attention, it focuses on relevant parts of an image while ignoring irrelevant details \cite{fan2021interpretability}. This approach addresses information overload and improves task efficiency. For example, Zhang et al. \cite{90} developed an XAI system to diagnose Helicobacter pylori (EADHI) infection using endoscopic features. The system combines ResNet50 and LSTM(long- and short-term memory) for feature extraction and employs gradient-boosting DT to analyze the contribution of gastric mucosal features to the diagnosis.}


\textbf{Proxy representation}: When the access object is unsuitable or cannot directly reference the target object, introducing a proxy object as a mediator between the access object and the target object can control access to the object. \textcolor{black}{This method is useful in complex optimization problems, where the solution space is large and computationally expensive.} In \cite{100}, they proposed a ML model that can provide reasonable reasons for the prediction and classify Parkinson's disease using SPECT DaTSCAN images.

\textbf{Knowledge distillation}: \textcolor{black}{Introduced by Hinton et al. \cite{hinton2015distilling}, this technique transfers knowledge from a complex teacher network to a simpler student network using soft targets. While effective, challenges remain in designing efficient teacher-student architectures.} In \cite{93}, an XAI model called the Bayesian Rule List (using rules and Bayesian analysis) is proposed. \textcolor{black}{It uses interpretable if-else statements to build accurate and interpretable predictive stroke models, advancing personalized medicine.}


\textbf{Feature extraction}: \textcolor{black}{This process converts data (such as text or images) into digitized features for ML, the feature is to enable computers to understand data better.} For example, Chaddad et al. \cite{chaddad2023texture} proposed a new radiomic feature called PCA-CNN, which uses principal component analysis (PCA) to extract principal features from various layers of CNN. \textcolor{black}{This approach was used to predict important indicators in patients with low-grade glioma (LGG) from MRI data.}


\begin{figure}[h] \centering
    \includegraphics[width=0.98\linewidth]{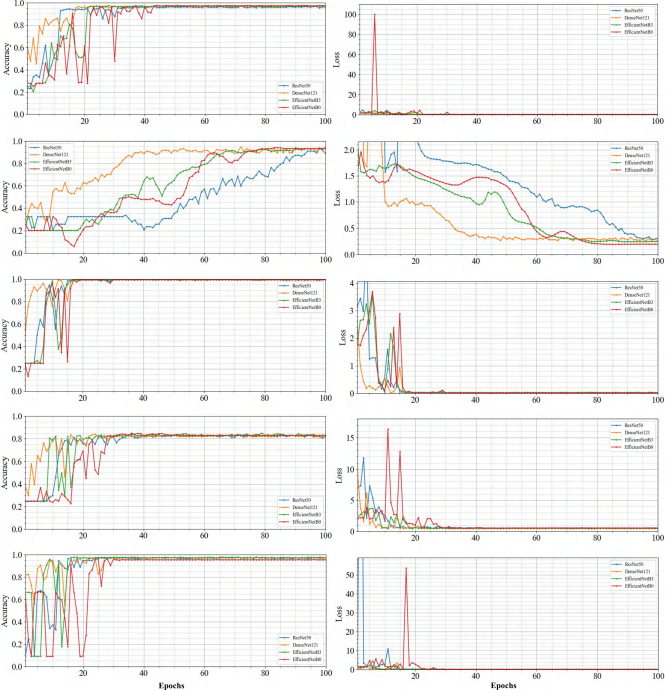}
    \caption{Performance comparison of ResNet50, DenseNet121, EfficientNetB3, and EfficientNetB0 (represent with lines of blue, orange, green, and red colors respectively) trained on different datasets. The five sets of graphs in each column represent the accuracy and loss of the same task using different models. The images in each row from top to bottom belong to the dataset of brain tumors, lung cancer, colon diseases, eye diseases, and Covid-19 diseases, respectively.}
    \label{fig:my_label 8}
\end{figure}

\section{Experiments} \label{5}
To comprehensively evaluate the performance of the three main XAI methods, Integrated Gradients (IG), Grad-CAM, and SHAP, we considered four public CNN models (ResNet50, DenseNet121, EfficientNetB3, and EfficientNetB0) for classification tasks. These models are applied to five medical imaging data sets (brain tumor, lung cancer, colon disease, eye disease, and Covid-19). It is important to mention that while there are other deep CNN models available, these four models are highly esteemed for medical image analysis due to their effectiveness and accuracy in handling complex image data. Through these trained CNN models, we are able to generate explanations for model decisions and evaluate these explanations using XAI methods. Our research objective is to explore the ability of these XAI methods to reveal the decision-making process of models, particularly how they explain the behavior of models in medical image classification tasks.

\begin{table}\scriptsize
    \setlength{\tabcolsep}{2.25mm}
    \caption{Summary of performance metrics for different models in classification tasks.}
    \begin{tabular}{ m{3cm} c c c c} 
     \toprule
        \textbf{Datasets} & \textbf{Accuracy} & \textbf{Precision} & \textbf{Recall} & \textbf{F1} \\
         \midrule
         \multicolumn{5}{c}{\textsc{ResNet50}} \\
         \midrule
           Brain tumor & 96.57 & 96.65 & 96.57 & 96.52  \\
           Lung Cancer& 78.73 &  82.30 & 78.73  & 78.05  \\
           Colon diseases& 99.00 & 99.03  &99.00  & 99.00  \\
           Eye diseases & 89.83 & 91.13 & 89.83 & 89.77  \\
           Covid-19 & 91.85 & 93.01 & 91.85 & 92.08  \\
           
            \midrule
          \multicolumn{5}{c}{\textsc{DenseNet121}} \\
          \midrule
           Brain tumor & 97.86 & 97.87 & 97.86 & 97.85  \\
           Lung Cancer& 81.90 & 83.82  & 81.90  &  81.97  \\
           Colon diseases& 99.62  & 99.63  & 99.62 & 99.62  \\
           Eye diseases & 89.72 & 91.46 &89.72  & 89.73  \\
           Covid-19 & 92.31 & 93.66 & 92.31 & 92.57  \\
           
          \midrule 
\multicolumn{5}{c}{\textsc{EfficientNet B3}} \\
         \midrule
           Brain tumor & 98.63 & 98.63 & 98.63 & 98.63 \\
           Lung Cancer & 88.25  & 88.97  & 88.25  &  88.23 \\
           Colon diseases& 99.25 & 99.25  &99.25  & 99.25   \\
        Eye diseases & 93.14  & 93.49 & 93.14 & 93.16\\
           Covid-19 diseases & 94.41 & 94.93 & 94.41 & 95.52  \\
          \midrule
\multicolumn{5}{c}{\textsc{EfficientNet B0}} \\
         \midrule
         Brain tumor & 98.78 & 98.78 & 98.78 &98.78 \\
        Lung Cancer& 85.08& 86.48  & 85.08 &  85.14 \\
           Colon diseases& 99.25 & 99.25  &99.25  & 99.25  \\
           Eye diseases & 93.62 & 93.85 &93.62  &93.62   \\
           Covid-19  & 93.09 & 93.53 & 93.09  & 93.21 \\

         \bottomrule
          \end{tabular}   
          \label{T:2}
    \end{table}

\begin{figure}[!ht] \centering
\includegraphics[width=0.98\linewidth]{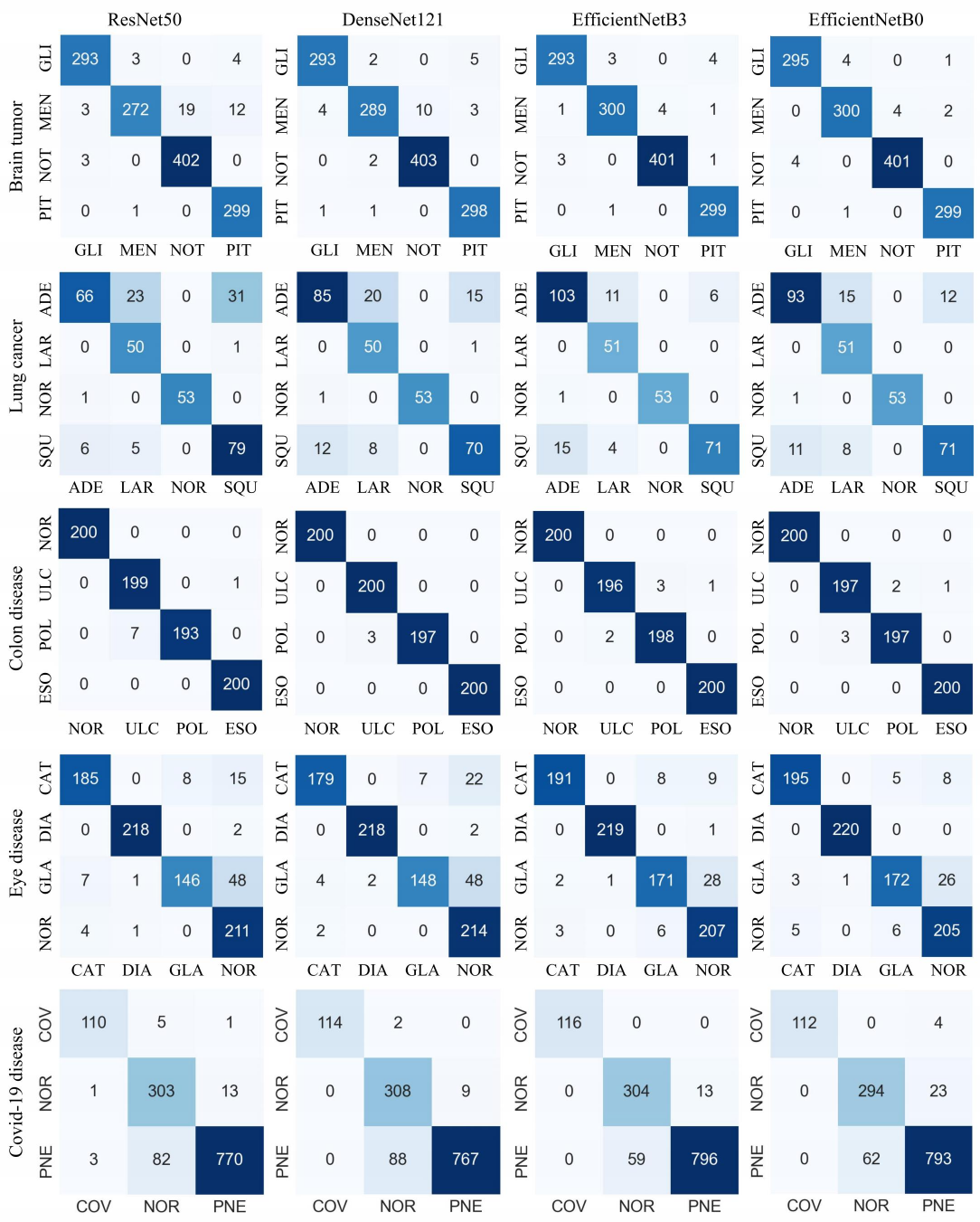}
    \caption{Confusion matrices overview. These matrices illustrate the accuracy of each model in distinguishing between categories, highlighting true and false predictions. The abbreviations in the figure are as follows: glioma tumors (GLI), meningioma tumors (MEN), no tumors (NOT) and pituitary tumors (PIT); adenocarcinoma (ADE), large cell carcinoma (LAR), normal cell (NOR) and squamous cell carcinoma (SQU); normal (NOR), ulcerative colitis (ULC), polyps (POL), and esophagitis (ESO); cataract (CAT), diabetic retinopathy (DIA), glaucoma (GLA) and normal (NOR); Covid-19 (COV), normal (NOR), and pneumonia (PNE).}
    \label{fig:my_label 15}
\end{figure}

\subsection{Hardware and software configuration}
This research used a hardware platform with advanced computing capabilities and ample memory to facilitate efficient training and assessment of DL models. The system features a 13th generation Intel® Core™ i9-13900K processor and is supported by 128GB of RAM. These specifications offer robust assistance to handle large data sets and perform computationally demanding operations. Graphics processing power is supported by the included NVIDIA GeForce RTX 4090 GPU, and its specialized large-capacity video memory significantly enhances the effectiveness of image processing and model training. Regarding the software setup, Python 3.9.18 is used as an experimental coding environment, allowing the accurate execution of models and analyses. The experimental configurations specified include the use of cross-entropy loss \cite{feng2021can}, a batch size of 32, 100 epochs, and a learning rate set at 0.001 for the Adam optimizer \cite{kingma2014adam} to ensure the stability and effectiveness of the training procedure.

\subsection{Dataset} We used five publicly available Kaggle data sets in the experiments. In the partition of data sets, for the brain tumor and Covid-19 datasets, which did not have a predefined validation set, we assign 20\% of the training set to serve as the validation set. In the case of the lung cancer and colon disease datasets, although a validation set was initially present, we incorporated it into the training set and subsequently repartitioned 20\% from this combined set to build a new validation set. Finally, for the eye disease dataset, we completely structured the division of the data, assigning 70\% to training, 10\% to validation and 20\% to tests to ensure a comprehensive evaluation framework. We present the number of data sets split as the following: n = total samples, tr = training samples, v = validation samples, and te = testing samples.

\textbf{Brain tumors} \cite{msoud_nickparvar_2021}: This data set consists of 7022 (tr=4571, v=1141, te=1311) MRI images of brain tumors, divided into four categories: glioma tumors (n=1621), meningioma tumors (n=1645), no tumors (n=2000) and pituitary tumors (n=1757). 

\textbf{Lung cancer} \cite{MOHAMED_HANY}: The data set includes four types of chest CT scan cancer images (n =1000, tr=581, v=94, te=315), namely adenocarcinoma (n=338, ADE), large cell carcinoma (n = 187, LCC), normal cell (n = 215, NOR) and squamous cell carcinoma (n=260, SCC). 

\textbf{Colon disease} \cite{FRANCIS_JESMAR_MONTALBO}: The data set comprises 6000 colon images (tr = 4160, v = 1040, te = 800) obtained by wireless capsule endoscopy, classified into normal, ulcerative colitis, polyps, and esophagitis, with 1500 images for each category.

\textbf{Eye diseases} \cite{GUNA_VENKAT_DODDI_2022}: This data set consists of four types of retinal images of eye disease: cataract (n = 1038), diabetic retinopathy (n = 1098), glaucoma (n = 1007) and normal (n = 1074), with a total of 4217 images (tr = 2698, v = 673, te = 846). 

\textbf{Covid-19 disease} \cite{PRASHANT_PATEL_2021}: The data set comprises 6432 CT images (tr = 4116, v = 1028, te = 1288) of Covid-19 cases grouped into three groups: Covid-19, normal, and pneumonia patients, with 576, 1583, 4274 images, respectively.

\begin{figure}[t] \centering
    \includegraphics[width=0.98\linewidth]{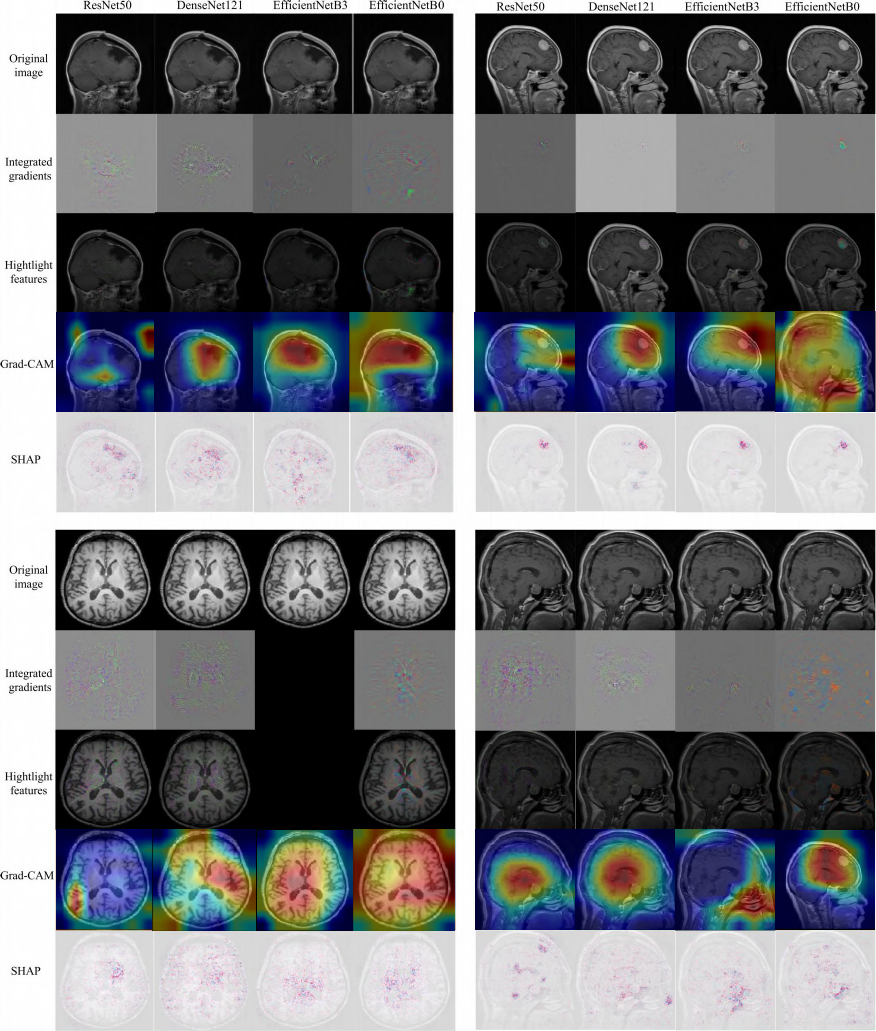}
    \caption{Example of XAI methods (Integrated Gradients, Grad-CAM, and SHAP) applied on brain tumor images. The classes of images are arranged from left to right and top to bottom as follows: glioma, meningioma, no tumors, and pituitary tumors.}
    \label{fig:my_label 9}
\end{figure}

\begin{figure}[h] \centering
    \includegraphics[width=0.98\linewidth]{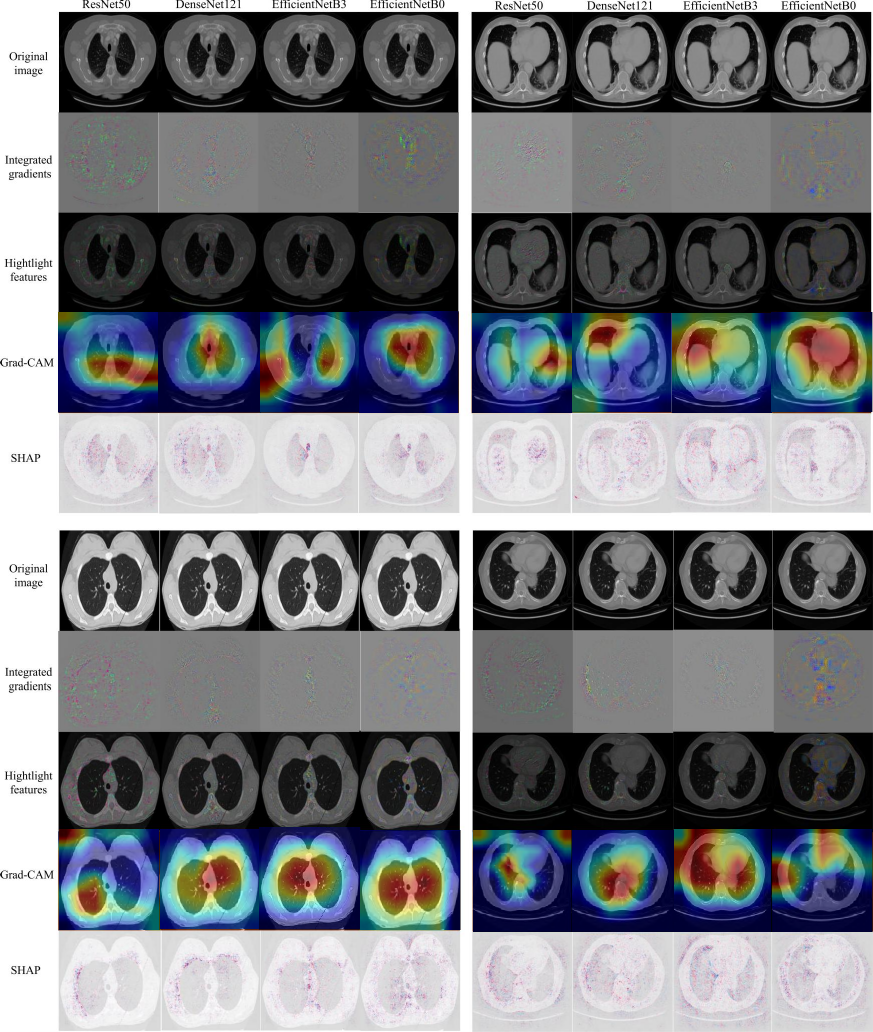}
    \caption{Applied Integrated Gradients, Grad-CAM, and SHAP in the lung cancer images. From left to right and from top to bottom, the classes of images are, respectively, adenocarcinoma, large cell carcinoma, normal cell, and squamous cell carcinoma.}
    \label{fig:my_label 10}
\end{figure}

\begin{figure}[h] \centering
    \includegraphics[width=0.98\linewidth]{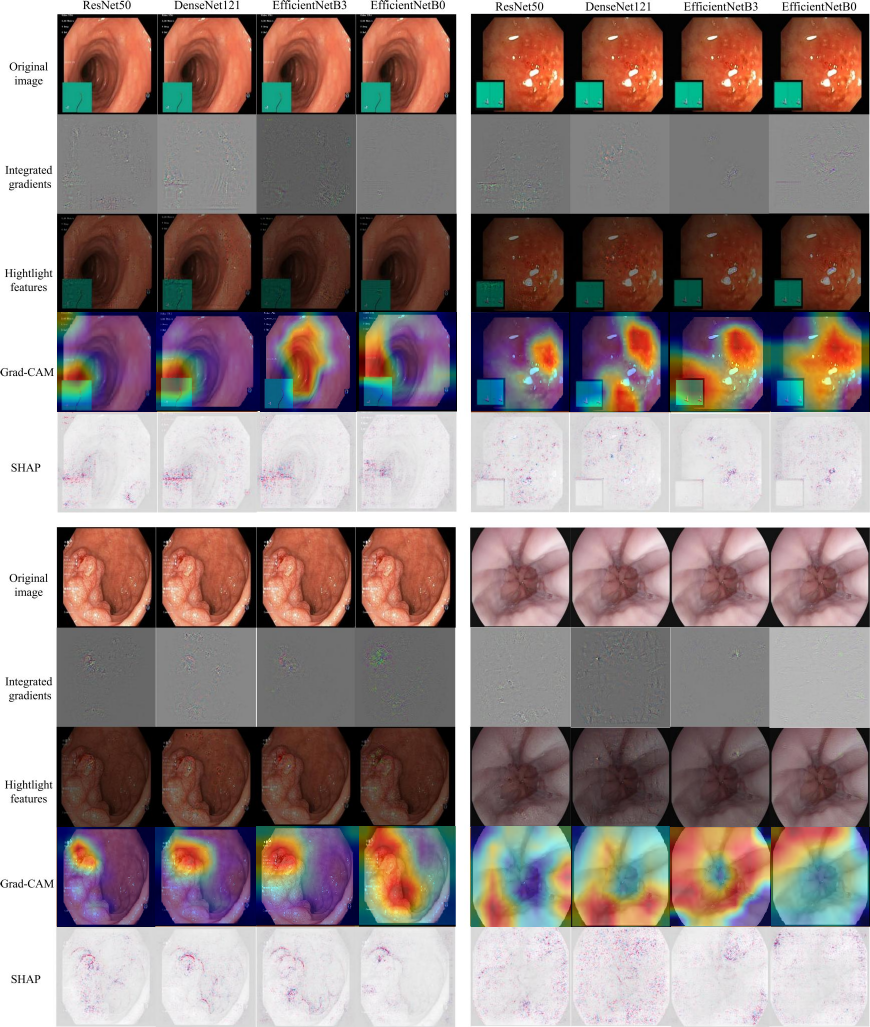}
    \caption{Applied Integrated Gradients, Grad-CAM, and SHAP in the colon diseases images. From left to right and top to bottom, the classes of images are respectively normal, ulcerative colitis, polyps, esophagitis.}
    \label{fig:my_label 11}
\end{figure}

\begin{figure}[h] \centering
    \includegraphics[width=0.98\linewidth]{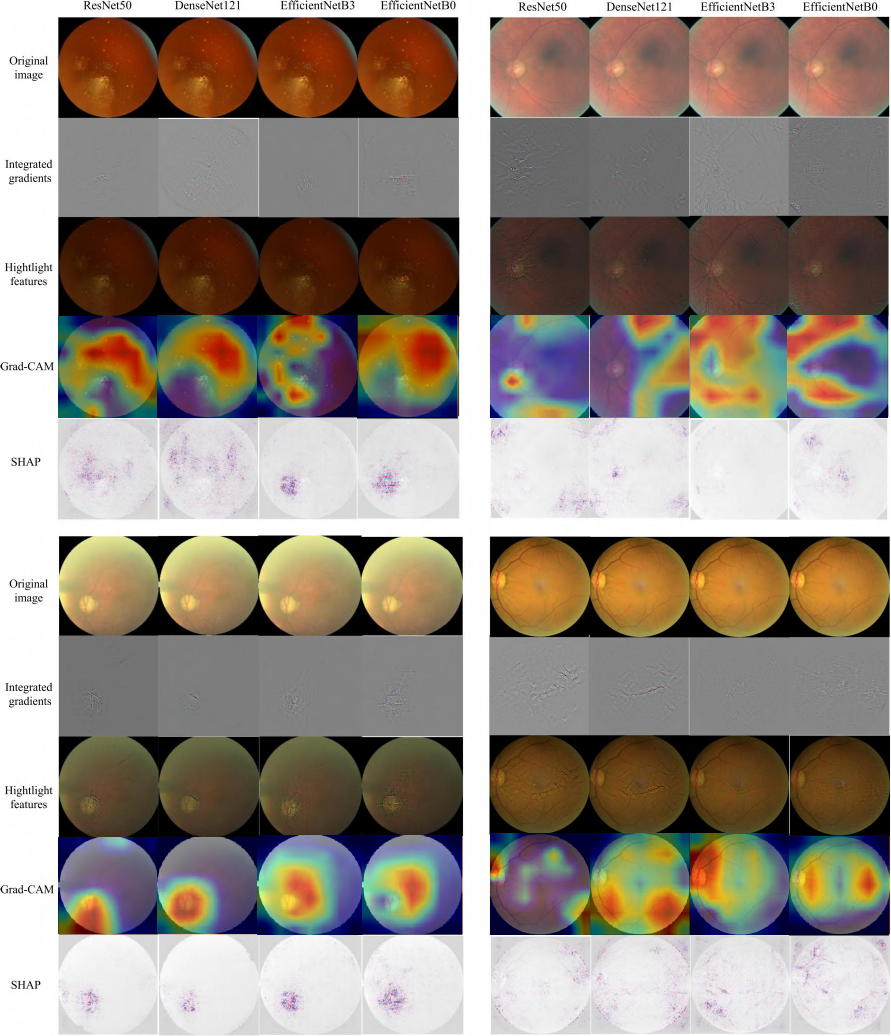}
    \caption{Example of XAI methods (Integrated Gradients, Grad-CAM, and SHAP) applied on eye diseases. The classes of images are arranged from left to right and top to bottom as follows: cataract, diabetic retinopathy, glaucoma, and normal.}
    \label{fig:my_label 12}
\end{figure}

\begin{figure*}\centering 
    \includegraphics[width=1\linewidth]{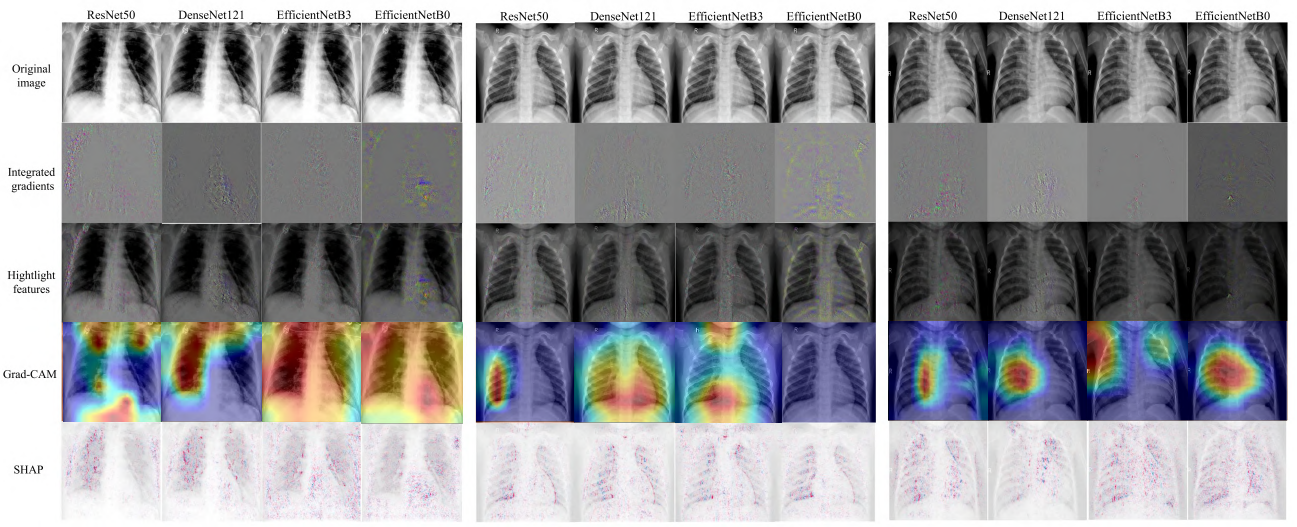}
    \caption{Example of XAI techniques: Integrated Gradients, Grad-CAM, and SHAP in images related to COVID-19 diseases. The image categories are Covid-19 (left), normal (middle) and pneumonia (right).}
    \label{fig:my_label 13}
\end{figure*}

\begin{figure}
    \centering
    \includegraphics[width=0.99\linewidth]{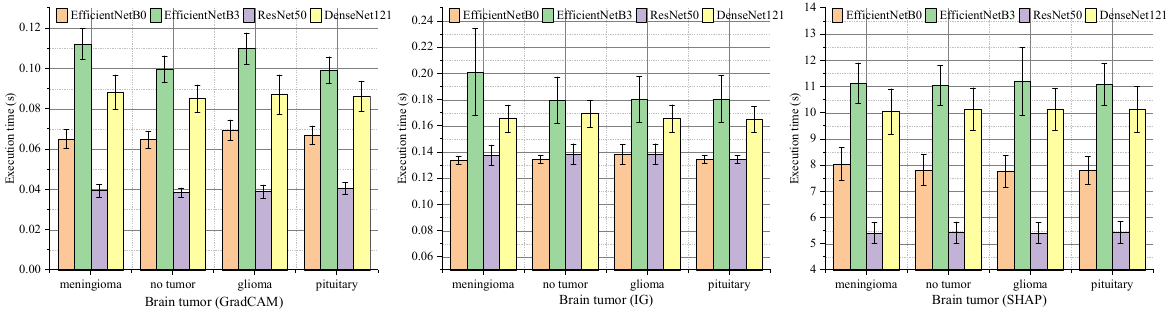}\\
    \includegraphics[width=0.99\linewidth]{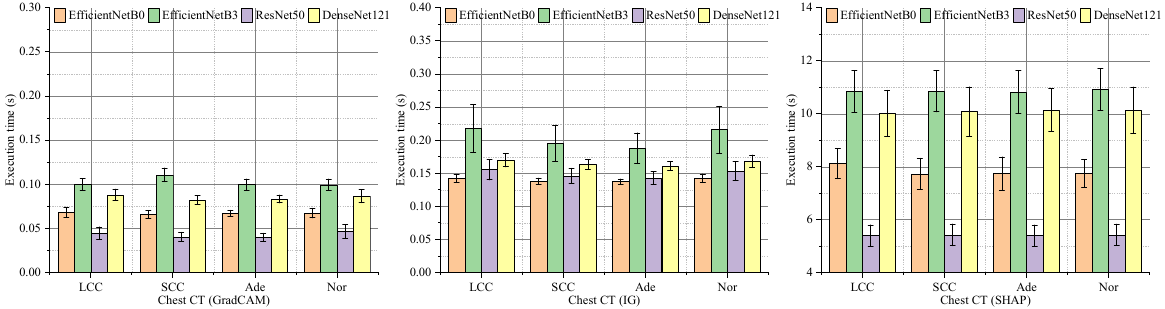}\\
    \includegraphics[width=0.99\linewidth]{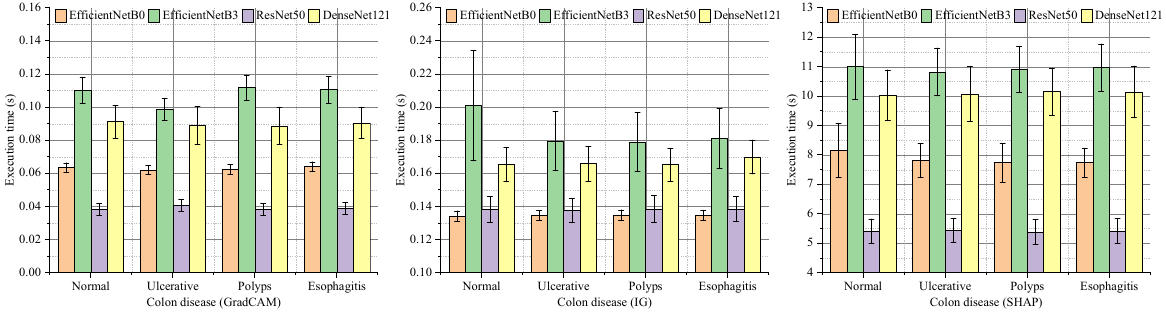}\\
    \includegraphics[width=0.99\linewidth]{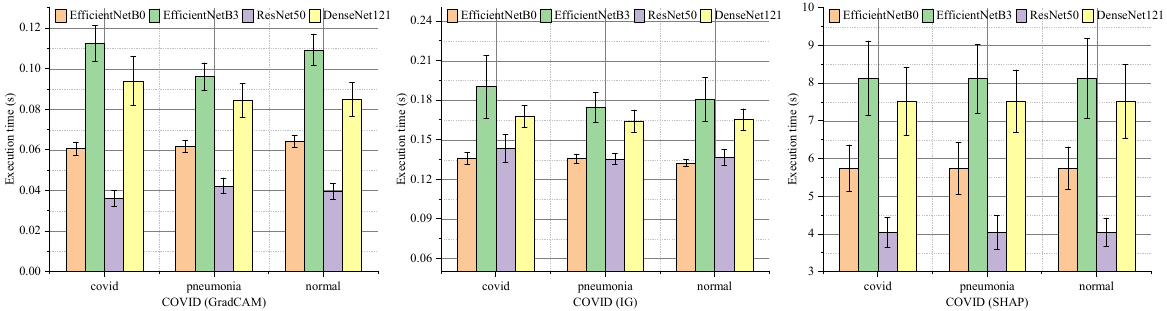}\\
    \includegraphics[width=0.99\linewidth]{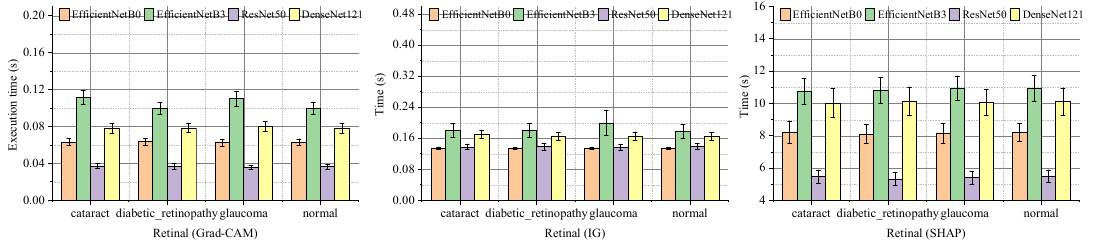}
    \caption{Execution time (Average with error bar) for Grad-CAM, IG and SHAP techniques on Brain tumor (\textbf{first row}), Chest CT (\textbf{second row}), Colon disease (\textbf{third row}), COVID-19 (\textbf{fourth row}) and Eye disease (\textbf{last row}) datasets.}
    \label{fig:Time_BT}
\end{figure}

\begin{figure}
    
    \centering
    \includegraphics[width=0.99\linewidth]{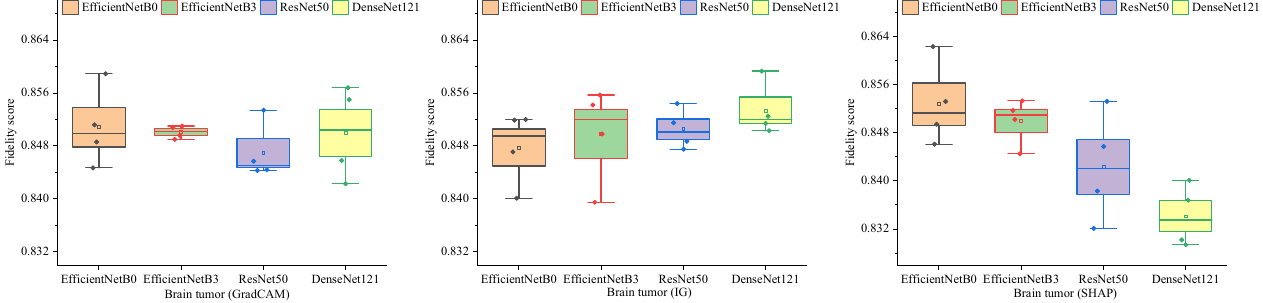}\\
    \includegraphics[width=0.99\linewidth]{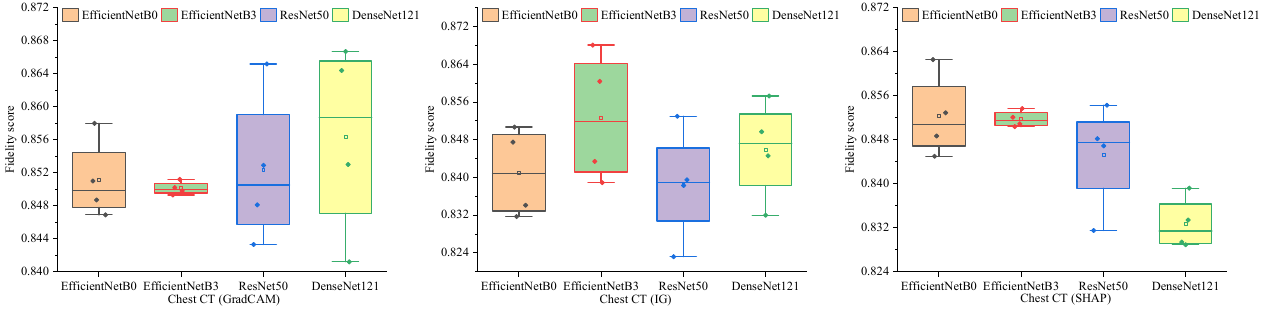}\\
    \includegraphics[width=0.99\linewidth]{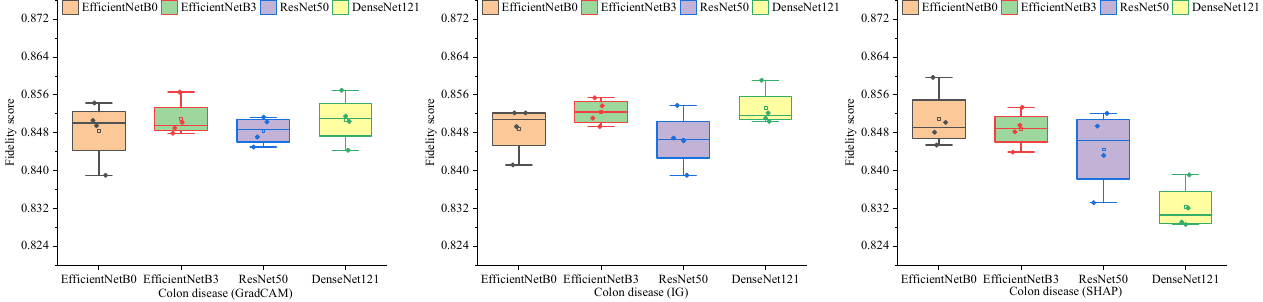}\\
    \includegraphics[width=0.99\linewidth]{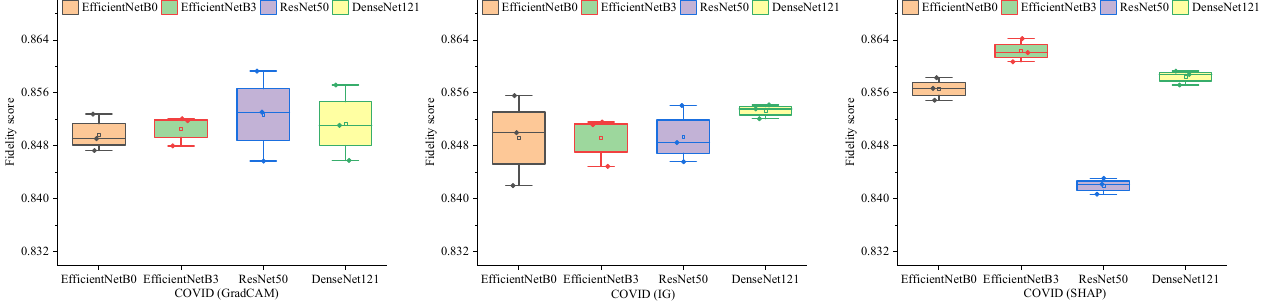}\\
    \includegraphics[width=0.99\linewidth]{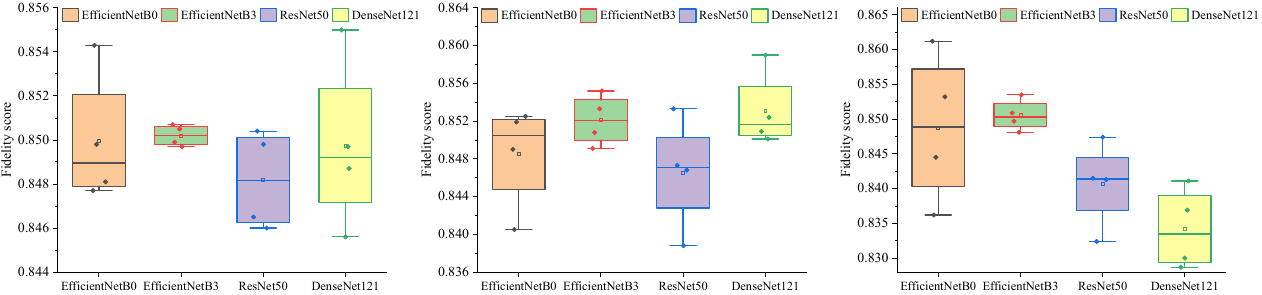}
    \vspace{3pt}
    \caption{Fidelity score for Grad-CAM, IG and SHAP techniques on Brain tumor (\textbf{first row}), Chest CT (\textbf{second row}), Colon disease (\textbf{third row}), COVID-19 (\textbf{fourth row}) and Eye disease (\textbf{last row}) datasets.}
    \label{fig:Time_Retinal}
\end{figure}

\subsection{Evaluation metrics} To assess the performance of the model, we use the following metrics: Precision, Recall, F1-score, Accuracy.

\begin{equation}
    Precision= \frac{T_{P}}{T_{P}+F_{P}}
\end{equation}   

\begin{equation}
    Recall=\frac{T_{P}}{T_{P}+F_{N}}
\end{equation}

\begin{equation}
    F1 score=2\times \frac{Precision\times Recall}{Precision+Recall}
\end{equation}

\begin{equation}
    Accuracy=\frac{T_{P}+T_{N}}{T_{P}+T_{N}+F_{P}+F_{N}}
\end{equation}

where $T_{P}$, $T_{N}$, $F_{P}$ and $F_{N}$ represent true positive, true negative, false positive, and false negative, respectively.

In addition, we introduce the fidelity score (F) to evaluate the trustworthiness of an explainability method and quantify the consistency between explanation and model prediction. It compares the prediction confidence of the model under normal conditions (original confidence) with the prediction confidence when adversarial perturbations are introduced. The closer the two values are, the more faithful the explanation method is to the behavior of the model.

\begin{equation}
    F = \frac{C_a}{C_o}, \quad 0 \leq F \leq 1
\end{equation}

where, $C_{o}$ represents original confidence of the model for a specific class (i.e., the predicted probability for the top class from the softmax output of the model). $C_{a}$ represents adversarial confidence, calculated as a perturbed version of $C_{o}$, with constraints ensuring 0 $\leq$ $C_{a}$ $\leq$ $C_{o}$. 

A high fidelity score (F close to 1) indicates a stable, robust, and reliable explanation method that aligns well with the model predictions. However, a low score (F close to 0) suggests poor alignment and robustness.

\subsection{Results} 

To summarize our simulation results, we present the performance metrics in the following form: [accuracy, precision, recall, F1 score, and number of correctly / incorrectly classified samples]. Table \ref{T:2} shows the performance metrics using the test data set. We also show the accuracy and loss of the validation sets across epochs for each model in Figure \ref{fig:my_label 8} and the confusion matrices in Figure \ref{fig:my_label 15}. We note that \textit{Integrated Gradients} maps demonstrate the impact of individual features in model decisions through the assignment of weights to pixels or features. However, \textit{Highlight Features} are images that have been enhanced by highlighting important features in the original image by weighting, which serves to make these important regions more visually prominent. 

\textit{Brain tumor}: EfficientNetB0 and EfficientNetB3 models show high performance metrics [98.63-98.78, 98.63-98.78, 98.63-98.78, 98.63-98.78, 1295/16-1293/18] with an accuracy of range value of $\geq$ 98\%. Figure \ref{fig:my_label 9} shows that integrated gradients and SHAP methods provide similar and accurate explanations, especially in identifying the meningioma tumor (upper right corner). However, the explanation of Grad-CAM shows poor generalization (e.g., identifies areas that are too large or too scattered).

\textit{Lung cancer}: EfficientNetB3 indicates high performance metrics [88.25, 88.97, 88.25, 88.23, 278/35]. The significant difference in accuracy may be due to the small size of the data set and insufficient feature learning. In addition, the imbalance of data classes is also a factor that affects the model performance. Figure \ref{fig:my_label 10} shows results similar to those of brain tumor images, where the explanation of Grad-CAM is relatively weak, indicating that it may not provide useful explanations in many cases.

\textit{Colon disease}: The relatively simple nature of the colon disease data set allows even deeper network structures to easily learn key features of different classes. For example, the EfficientNet-B3 shows high performance metrics [99.25, 99.25, 99.25, 99.25, 794/6]. Figure \ref{fig:my_label 11} shows that the explanations provided by different XAI methods are largely consistent, further confirming the stability of the model in the identification of key features.

\textit{Eye disease}: EfficientNetB0 performed best with high performance metrics [93.62, 93.85, 93.62, 93.62, 792/54], slightly higher than 93.14\% of EfficientNetB3. Figure \ref{fig:my_label 12} shows that Grad-CAM results may consider a large region of interest, while Integrated Gradients and SHAP provide similar explanations.

\textit{Covid-19 diseases}: EfficientNetB3 still achieved the best performance, with high performance metrics [94.41, 94.93, 94.41, 95.52, 1216/72]. The reason for the slight difference between EfficientNetB3 and EfficientNetB0 may be that, in this dataset, EfficientNetB3 demonstrates a higher classification accuracy through its larger number of parameters and a deeper network architecture. In Figure \ref{fig:my_label 13}, it can be observed that SHAP and Integrated Gradients offer comparable, yet restricted explanations, while Grad-CAM demonstrates a higher explanation in certain images, but lacks clarity in others.

Further analysis of how classification model performance affects the effectiveness of XAI methods shows a clear correlation between model accuracy and the quality of feature explanations. For data sets such as brain tumors and colon disease, where the models achieve high accuracy, XAI methods are able to focus on and highlight key regions, such as tumors or lesions, with high precision. These data sets show lower misinterpretation rates, with more concentrated and consistent explanations. In contrast, for the lung cancer dataset, where the model accuracy is lower (78-88\%), the performance of XAI methods is less reliable. Specifically, Grad-CAM tends to produce larger, less focused regions in its explanations. This highlights that improving model performance improves classification accuracy and improves the precision and consistency of XAI methods, ultimately making their use in medical image classification tasks more trustworthy and effective.

\noindent\textbf{Quantitative analysis.} Figures \ref{fig:Time_BT} and \ref{fig:Time_Retinal} illustrate the execution time and the Fidelity score using GradCAM, IG, and SHAP on Brain tumor, Chest CT, Colon disease, COVID-19, and Eye disease datasets. For a comprehensive evaluation of these XAI techniques, we provide results based on the average $\pm$ standard deviation of the execution time/Fidelity score of all test images. For example, the use of GradCAM results in a more stable Fidelity score compared to IG and SHAP (e.g., the Fidelity score of SHAP decreases by about 0.01 using DenseNet121 compared to EfficientNetB0), highlighting its consistency across different deep models. In addition, the execution time of GradCAM and IG is less than that of SHAP (e.g., approximately 0.04 seconds using ResNet50 for GradCAM and approximately 5.5 seconds using ResNet50 for SHAP). These results are similar for Colon disease and Eye disease datasets. However, for the COVID-19 dataset, SHAP indicates a higher Fidelity score using EfficientNetB0 and EfficientNetB3 compared to GradCAM and IG (e.g., greater than 0.856). This suggests that SHAP can highlight key features in COVID-19 images. In summary, these findings suggest that GradCAM and IG can provide feasible interpretability with acceptable computational costs, which is of great value for low-resource devices \cite{dwivedi2024efficient}.

\subsection{Evaluation methods} 
We conducted numerous simulation experiments on medical image classification to demonstrate the impact of XAI methods. However, there are numerous methods available to evaluate the impact of XAI methods as follows.

\textit{Quantitative measurement}: The effectiveness of XAI methods can be evaluated through a series of quantitative indicators, such as the accuracy of the explanation (whether the explanation accurately maps the decision-making process of the model), coverage (whether the explanation covers all important decision-making factors of the model), and subjective evaluation by the user of the accuracy of the explanation \cite{nauta2023anecdotal}.

\textit{User-centric research}: Such studies facilitate the collection of data on the levels of interpretability, utility, and satisfaction among users. Using surveys, in-depth interviews, or participatory experiments, this methodology aims to gauge the perceptions and valuations of non-technical and expert users of the explanations provided \cite{haque2023explainable}.

\textit{Comparative analysis}: The efficacy of XAI methodologies can be assessed by juxtaposing their output with established benchmarks or expert analyzes. Such comparative analyzes are important to highlight the discrepancies between XAI-generated explanations and those derived from human expertise \cite{alizadehsani2024explainable}.

\textit{Visualization techniques}: Leveraging visualization tools and methodologies enables the demonstration and critical evaluation of the explanations obtained by XAI methods. Effective visualization helps to make the decision of the model and the content of the explanations more accessible and comprehensible to users \cite{hassija2024interpreting}.

\textit{Multi-dimensional evaluation}: This approach advocates for a holistic examination across various dimensions, including transparency of explanations, user trust, the efficacy of the model, and the influence of explanations on decision-making processes, providing a rounded evaluative perspective \cite{lipton2018mythos}. Using these evaluative strategies allows an in-depth discernment of the multifaceted impacts and the inherent value of XAI techniques. Although the direct application of these evaluation measures was beyond the scope of this study, indirect insights gleaned from experimental observations underscore the transformative potential of XAI to improve model transparency and foster user trust. These methodologies elucidate the pathways for future inquiries, highlighting the importance of integrating XAI within the realms of disease classification.

\section{Challenges of XAI} \label{6}

We introduce the challenges faced by XAI, focusing on the medical field, and summarize these challenges in Table \ref{T:3}.

\begin{table*}\footnotesize
    \setlength{\tabcolsep}{3mm}
    \caption{XAI challenges in healthcare in recent two years.}
    \begin{tabular}{m{0.5cm}  m{3.7cm} m{12cm}} 
     \toprule
        \textbf{Ref} & \textbf{Main challenges} & \textbf{Description} \\
         \midrule
          \cite{105} & Ethical issues & Responsibility ethics determine who will be responsible for decisions that affect patient health\\
          \cite{105} & Legal issues & Strengthen laws and regulations to ensure the security of medical data privacy\\
          \cite{105} & Robustness issues & The validated applications in the health field based on AI or ML are still immature and early\\
          \cite{106} & Transparency & Lack of accessibility to models is also one of the obstacles to the application of AI in the medical field  \\
          \cite{106} & Explainability & The decisions made by AI need be understandable to doctors and patients \\
          \cite{107} & The definition of a ground truth & To ensure that the decisions made by clinical doctors and AI are basically consistent \\

          \cite{108} & Privacy and legal issues & Medical data leakage can cause mental harm to patients, and law is the best protection measure \\
          \cite{108} & Trust issues and explainability & When the 'black box' model is used for major or even fatal diseases, patients may develop distrust of it\\
          \cite{109} & Understandability & Explanations should enable patients without professional knowledge to understand \\
          \cite{109} & Truthfulness & Explanation are able to fully and accurately reflect the decision and its process \\
          \cite{109} &  Computational efficiency & The time for providing explanations are reasonable and fast \\
          \cite{viswan2024explainable} & Domain experts' input & Many researchers may lack medical expertise, leading to errors in assessing the quality of an explanation.\\
          
         \bottomrule
          \end{tabular}   
          \label{T:3}
\end{table*}

\textbf{Transparency:} The primary function of XAI is to make the model transparent and to allow the end users to understand the output of the model. Therefore, transparency is also the biggest challenge for XAI. For example, when XAI increases the transparency of the system, many of the challenges faced by XAI also decrease, including a deeper understanding of user output (understandability issues), which dramatically improves user experience and trust in system decisions (trust challenge), thus strengthening the interaction between users and the system \cite{mohseni2021multidisciplinary}. Nevertheless, transparency and predictive ability are usually not a trade-off, so it is not an excellent choice to blindly improve the transparency of the model \cite{ueda2024fairness}.

\textbf{Standardization and normalization:} As XAI is a relatively young field, there is no standard term in the community. For example, "explainability" and "explicability", "feature importance", and "feature relevance" refer to the same concept. The standard and formal principles for researchers in developing and designing XAI systems and solutions are critical; a universal and unified theory is needed to approximate the structure and intent of interpretation \cite{mohseni2021multidisciplinary}.

\textbf{Robustness issues:} Systems that exhibit high robustness are capable of handling significant levels of interference or noise. They are able to efficiently avert system failures or incorrect output outcomes when errors occur. For XAI systems, the robustness is recommended to be relatively high. For example, if the input being interpreted undergoes changes that are not significant enough to impact the predicted outcomes, the explanation will remain relatively unchanged \cite{vilone2021notions}.

\textbf{Privacy:} User privacy is the most common carrier of information in social networks. Without the influence of privacy protection measures, the communication scheme will be difficult to directly apply to the actual environment, and the content captured and stored in the internal representation of the model cannot be understood, which can lead to disclosure of privacy \cite{patra2024xai}. Therefore, it is necessary to ensure that the XAI algorithm does not threaten the privacy of the data.

\textbf{Explainable security:} With the continuous progress of the digital world, systems and software are generally vulnerable to a wide range of security threats and network attacks. Explainable security is needed because all different stakeholders in the system want it to be secure (except attackers) \cite{34}. 

\textbf{Evaluation:} Psychologists and social scientists have studied how humans evaluate interpretations. Within their discipline, interpretation evaluation refers to the process used by the interpreter to determine whether the interpretation is satisfactory \cite{35}. Evaluation can be divided into two models: 1) evaluation and 2) interpretation evaluation. For the evaluation, we focus on the generalization of the system and evaluate the quality of the prediction. For interpretative evaluation, we focus on predictability, faithfulness, and persuasion of interpretation, and evaluate the quality of interpretation \cite{36}.

\section{Conclusion} \label{7}
This study, through a recent review of the literature, clarifies the classification methods of XAI and its prospects for application in healthcare, especially highlighting its practical value in computer-aided diagnosis. By applying various CNN classification models with XAI techniques to common public medical datasets, this research validates the potential of XAI to enhance diagnostic accuracy and points towards future research directions, namely improving the adaptability, explainability, and practical application of XAI methods. In facing the challenges of improving the explainability and user understanding in XAI, our future research will focus on refining and optimizing explanation models and exploring more intuitive and interactive explanation interfaces, enabling users to understand and trust the AI decision-making process more easily.

{\footnotesize
\bibliographystyle{IEEEtran}
\bibliography{ref}

\begin{thebibliography}{100}
\providecommand{\url}[1]{#1}
\csname url@samestyle\endcsname
\providecommand{\newblock}{\relax}
\providecommand{\bibinfo}[2]{#2}
\providecommand{\BIBentrySTDinterwordspacing}{\spaceskip=0pt\relax}
\providecommand{\BIBentryALTinterwordstretchfactor}{4}
\providecommand{\BIBentryALTinterwordspacing}{\spaceskip=\fontdimen2\font plus
\BIBentryALTinterwordstretchfactor\fontdimen3\font minus \fontdimen4\font\relax}
\providecommand{\BIBforeignlanguage}[2]{{%
\expandafter\ifx\csname l@#1\endcsname\relax
\typeout{** WARNING: IEEEtran.bst: No hyphenation pattern has been}%
\typeout{** loaded for the language `#1'. Using the pattern for}%
\typeout{** the default language instead.}%
\else
\language=\csname l@#1\endcsname
\fi
#2}}
\providecommand{\BIBdecl}{\relax}
\BIBdecl

\bibitem{chaddad2025eamapg}
A.~Chaddad, Y.~Jiang, T.~S. Daqqaq, and R.~Kateb, ``Eamapg: Explainable adversarial model analysis via projected gradient descent,'' \emph{Computers in Biology and Medicine}, vol. 188, p. 109788, 2025.

\bibitem{hassan2023explaining}
M.~M. Hassan, S.~A. AlQahtani, A.~Alelaiwi, and J.~P. Papa, ``Explaining covid-19 diagnosis with taylor decompositions,'' \emph{Neural Computing and Applications}, vol.~35, no.~30, pp. 22\,087--22\,100, 2023.

\bibitem{sivamohan2023optimized}
S.~Sivamohan and S.~Sridhar, ``An optimized model for network intrusion detection systems in industry 4.0 using xai based bi-lstm framework,'' \emph{Neural Computing and Applications}, vol.~35, no.~15, pp. 11\,459--11\,475, 2023.

\bibitem{he2016deep}
K.~He, X.~Zhang, S.~Ren, and J.~Sun, ``Deep residual learning for image recognition,'' in \emph{Proceedings of the IEEE conference on computer vision and pattern recognition}, 2016, pp. 770--778.

\bibitem{krizhevsky2012imagenet}
A.~Krizhevsky, I.~Sutskever, and G.~E. Hinton, ``Imagenet classification with deep convolutional neural networks,'' \emph{Advances in neural information processing systems}, vol.~25, 2012.

\bibitem{simonyan2014very}
K.~Simonyan and A.~Zisserman, ``Very deep convolutional networks for large-scale image recognition,'' \emph{arXiv preprint arXiv:1409.1556}, 2014.

\bibitem{han2022survey}
K.~Han, Y.~Wang, H.~Chen, X.~Chen, J.~Guo, Z.~Liu, Y.~Tang, A.~Xiao, C.~Xu, Y.~Xu \emph{et~al.}, ``A survey on vision transformer,'' \emph{IEEE transactions on pattern analysis and machine intelligence}, vol.~45, no.~1, pp. 87--110, 2022.

\bibitem{radford2021learning}
A.~Radford, J.~W. Kim, C.~Hallacy, A.~Ramesh, G.~Goh, S.~Agarwal, G.~Sastry, A.~Askell, P.~Mishkin, J.~Clark \emph{et~al.}, ``Learning transferable visual models from natural language supervision,'' in \emph{International conference on machine learning}.\hskip 1em plus 0.5em minus 0.4em\relax PMLR, 2021, pp. 8748--8763.

\bibitem{arrieta2020explainable}
A.~B. Arrieta, N.~D{\'\i}az-Rodr{\'\i}guez, J.~Del~Ser, A.~Bennetot, S.~Tabik, A.~Barbado, S.~Garc{\'\i}a, S.~Gil-L{\'o}pez, D.~Molina, R.~Benjamins \emph{et~al.}, ``Explainable artificial intelligence (xai): Concepts, taxonomies, opportunities and challenges toward responsible ai,'' \emph{Information fusion}, vol.~58, pp. 82--115, 2020.

\bibitem{dwivedi2023explainable}
R.~Dwivedi, D.~Dave, H.~Naik, S.~Singhal, R.~Omer, P.~Patel, B.~Qian, Z.~Wen, T.~Shah, G.~Morgan \emph{et~al.}, ``Explainable ai (xai): Core ideas, techniques, and solutions,'' \emph{ACM Computing Surveys}, vol.~55, no.~9, pp. 1--33, 2023.

\bibitem{dhar2023challenges}
T.~Dhar, N.~Dey, S.~Borra, and R.~S. Sherratt, ``Challenges of deep learning in medical image analysis—improving explainability and trust,'' \emph{IEEE Transactions on Technology and Society}, vol.~4, no.~1, pp. 68--75, 2023.

\bibitem{hassija2024interpreting}
V.~Hassija, V.~Chamola, A.~Mahapatra, A.~Singal, D.~Goel, K.~Huang, S.~Scardapane, I.~Spinelli, M.~Mahmud, and A.~Hussain, ``Interpreting black-box models: a review on explainable artificial intelligence,'' \emph{Cognitive Computation}, vol.~16, no.~1, pp. 45--74, 2024.

\bibitem{el2023utilizing}
N.~El-Rashidy, N.~E. ElSayed, A.~El-Ghamry, and F.~M. Talaat, ``Utilizing fog computing and explainable deep learning techniques for gestational diabetes prediction,'' \emph{Neural Computing and Applications}, vol.~35, no.~10, pp. 7423--7442, 2023.

\bibitem{jiang2025knowledge}
Y.~Jiang, X.~Zhao, Y.~Wu, and A.~Chaddad, ``A knowledge distillation-based approach to enhance transparency of classifier models,'' \emph{arXiv preprint arXiv:2502.15959}, 2025.

\bibitem{chaddad2023survey}
A.~Chaddad, J.~Peng, J.~Xu, and A.~Bouridane, ``Survey of explainable ai techniques in healthcare,'' \emph{Sensors}, vol.~23, no.~2, p. 634, 2023.

\bibitem{chaddad2023enhancing}
A.~Chaddad and Y.~Wu, ``Enhancing classification tasks through domain adaptation strategies,'' in \emph{2023 IEEE International Conference on Bioinformatics and Biomedicine (BIBM)}.\hskip 1em plus 0.5em minus 0.4em\relax IEEE, 2023, pp. 1832--1835.

\bibitem{sindiramutty2024modern}
S.~R. Sindiramutty, C.~E. Tan, W.~J. Tee, S.~P. Lau, S.~Balakrishnan, S.~Kaur, H.~Jazri, and M.~F.~A. Aslam, ``Modern smart cities and open research challenges and issues of explainable artificial intelligence,'' \emph{Advances in Explainable AI Applications for Smart Cities}, pp. 389--424, 2024.

\bibitem{sharma2024explainable}
B.~Sharma, L.~Sharma, C.~Lal, and S.~Roy, ``Explainable artificial intelligence for intrusion detection in iot networks: A deep learning based approach,'' \emph{Expert Systems with Applications}, vol. 238, p. 121751, 2024.

\bibitem{arreche2024xai}
O.~Arreche, T.~R. Guntur, J.~W. Roberts, and M.~Abdallah, ``E-xai: Evaluating black-box explainable ai frameworks for network intrusion detection,'' \emph{IEEE Access}, 2024.

\bibitem{patwary2024explainable}
A.~L. Patwary and A.~J. Khattak, ``Explainable artificial intelligence for decarbonization: Alternative fuel vehicle adoption in disadvantaged communities,'' \emph{International Journal of Sustainable Transportation}, pp. 1--15, 2024.

\bibitem{wen2023use}
B.~Wen and A.~Chaddad, ``The use of explainable artificial intelligence in medicine,'' in \emph{2023 IEEE International Conference on E-health Networking, Application \& Services (Healthcom)}.\hskip 1em plus 0.5em minus 0.4em\relax IEEE, 2023, pp. 251--252.

\bibitem{ghnemat2023explainable}
R.~Ghnemat, S.~Alodibat, and Q.~Abu Al-Haija, ``Explainable artificial intelligence (xai) for deep learning based medical imaging classification,'' \emph{Journal of Imaging}, vol.~9, no.~9, p. 177, 2023.

\bibitem{farrag2023explainable}
A.~Farrag, G.~Gad, Z.~M. Fadlullah, M.~M. Fouda, and M.~Alsabaan, ``An explainable ai system for medical image segmentation with preserved local resolution: Mammogram tumor segmentation,'' \emph{IEEE Access}, 2023.

\bibitem{van2022explainable}
B.~H. Van~der Velden, H.~J. Kuijf, K.~G. Gilhuijs, and M.~A. Viergever, ``Explainable artificial intelligence (xai) in deep learning-based medical image analysis,'' \emph{Medical Image Analysis}, vol.~79, p. 102470, 2022.

\bibitem{chaddad2024generalizable}
A.~Chaddad, Y.~Hu, Y.~Wu, B.~Wen, and R.~Kateb, ``Generalizable and explainable deep learning for medical image computing: An overview,'' \emph{Current Opinion in Biomedical Engineering}, p. 100567, 2024.

\bibitem{hossain2023explainable}
M.~I. Hossain, G.~Zamzmi, P.~R. Mouton, M.~S. Salekin, Y.~Sun, and D.~Goldgof, ``Explainable ai for medical data: Current methods, limitations, and future directions,'' \emph{ACM Computing Surveys}, 2023.

\bibitem{32}
S.~R. Islam, W.~Eberle, S.~K. Ghafoor, and M.~Ahmed, ``Explainable artificial intelligence approaches: A survey,'' \emph{arXiv preprint arXiv:2101.09429}, 2021.

\bibitem{samek2017explainable}
W.~Samek, ``Explainable artificial intelligence: Understanding, visualizing and interpreting deep learning models,'' \emph{arXiv preprint arXiv:1708.08296}, 2017.

\bibitem{singh2020explainable}
A.~Singh, S.~Sengupta, and V.~Lakshminarayanan, ``Explainable deep learning models in medical image analysis,'' \emph{Journal of imaging}, vol.~6, no.~6, p.~52, 2020.

\bibitem{el2024exhyptnet}
E.-S.~A. El-Dahshan, M.~M. Bassiouni, S.~K. Khare, R.-S. Tan, and U.~R. Acharya, ``Exhyptnet: An explainable diagnosis of hypertension using efficientnet with ppg signals,'' \emph{Expert Systems with Applications}, vol. 239, p. 122388, 2024.

\bibitem{chen2024deep}
T.-C.~T. Chen, H.-C. Wu, and M.-C. Chiu, ``A deep neural network with modified random forest incremental interpretation approach for diagnosing diabetes in smart healthcare,'' \emph{Applied Soft Computing}, vol. 152, p. 111183, 2024.

\bibitem{lalithadevi2023feasibility}
B.~Lalithadevi, S.~Krishnaveni, and J.~S.~C. Gnanadurai, ``A feasibility study of diabetic retinopathy detection in type ii diabetic patients based on explainable artificial intelligence,'' \emph{Journal of Medical Systems}, vol.~47, no.~1, p.~85, 2023.

\bibitem{dharmarathne2024novel}
G.~Dharmarathne, T.~N. Jayasinghe, M.~Bogahawaththa, D.~Meddage, and U.~Rathnayake, ``A novel machine learning approach for diagnosing diabetes with a self-explainable interface,'' \emph{Healthcare Analytics}, vol.~5, p. 100301, 2024.

\bibitem{wani2024deepxplainer}
N.~A. Wani, R.~Kumar, and J.~Bedi, ``Deepxplainer: An interpretable deep learning based approach for lung cancer detection using explainable artificial intelligence,'' \emph{Computer Methods and Programs in Biomedicine}, vol. 243, p. 107879, 2024.

\bibitem{khanna2023machine}
V.~V. Khanna, K.~Chadaga, N.~Sampathila, S.~Prabhu, and R.~Chadaga, ``A machine learning and explainable artificial intelligence triage-prediction system for covid-19,'' \emph{Decision Analytics Journal}, p. 100246, 2023.

\bibitem{hossain2023cardiovascular}
M.~M. Hossain, M.~S. Ali, M.~M. Ahmed, M.~R.~H. Rakib, M.~A. Kona, S.~Afrin, M.~K. Islam, M.~M. Ahsan, S.~M. R.~H. Raj, and M.~H. Rahman, ``Cardiovascular disease identification using a hybrid cnn-lstm model with explainable ai,'' \emph{Informatics in Medicine Unlocked}, vol.~42, p. 101370, 2023.

\bibitem{mahim2024unlocking}
S.~Mahim, M.~S. Ali, M.~O. Hasan, A.~A.~N. Nafi, A.~Sadat, S.~Al~Hasan, B.~Shareef, M.~M. Ahsan, M.~K. Islam, M.~S. Miah \emph{et~al.}, ``Unlocking the potential of xai for improved alzheimer’s disease detection and classification using a vit-gru model,'' \emph{IEEE Access}, 2024.

\bibitem{nafisah2024tuberculosis}
S.~I. Nafisah and G.~Muhammad, ``Tuberculosis detection in chest radiograph using convolutional neural network architecture and explainable artificial intelligence,'' \emph{Neural Computing and Applications}, vol.~36, no.~1, pp. 111--131, 2024.

\bibitem{7}
S.~N. Payrovnaziri, Z.~Chen, P.~Rengifo-Moreno, T.~Miller, J.~Bian, J.~H. Chen, X.~Liu, and Z.~He, ``Explainable artificial intelligence models using real-world electronic health record data: a systematic scoping review,'' \emph{Journal of the American Medical Informatics Association}, vol.~27, no.~7, pp. 1173--1185, 2020.

\bibitem{fontes2024application}
M.~Fontes, J.~D.~S. De~Almeida, and A.~Cunha, ``Application of example-based explainable artificial intelligence (xai) for analysis and interpretation of medical imaging: a systematic review,'' \emph{IEEE Access}, vol.~12, pp. 26\,419--26\,427, 2024.

\bibitem{31}
W.~Ding, M.~Abdel-Basset, H.~Hawash, and A.~M. Ali, ``Explainability of artificial intelligence methods, applications and challenges: A comprehensive survey,'' \emph{Information Sciences}, 2022.

\bibitem{selvaraju2017grad}
R.~R. Selvaraju, M.~Cogswell, A.~Das, R.~Vedantam, D.~Parikh, and D.~Batra, ``Grad-cam: Visual explanations from deep networks via gradient-based localization,'' in \emph{Proceedings of the IEEE international conference on computer vision}, 2017, pp. 618--626.

\bibitem{10085971}
A.~Chaddad, Q.~Lu, J.~Li, Y.~Katib, R.~Kateb, C.~Tanougast, A.~Bouridane, and A.~Abdulkadir, ``Explainable, domain-adaptive, and federated artificial intelligence in medicine,'' \emph{IEEE/CAA Journal of Automatica Sinica}, vol.~10, no.~4, pp. 859--876, 2023.

\bibitem{katzmann2021explaining}
A.~Katzmann, O.~Taubmann, S.~Ahmad, A.~M{\"u}hlberg, M.~S{\"u}hling, and H.-M. Gro{\ss}, ``Explaining clinical decision support systems in medical imaging using cycle-consistent activation maximization,'' \emph{Neurocomputing}, vol. 458, pp. 141--156, 2021.

\bibitem{17}
A.~B. Arrieta, N.~D{\'\i}az-Rodr{\'\i}guez, J.~Del~Ser, A.~Bennetot, S.~Tabik, A.~Barbado, S.~Garc{\'\i}a, S.~Gil-L{\'o}pez, D.~Molina, R.~Benjamins \emph{et~al.}, ``Explainable artificial intelligence (xai): Concepts, taxonomies, opportunities and challenges toward responsible ai,'' \emph{Information fusion}, vol.~58, pp. 82--115, 2020.

\bibitem{antwarg2021explaining}
L.~Antwarg, R.~M. Miller, B.~Shapira, and L.~Rokach, ``Explaining anomalies detected by autoencoders using shapley additive explanations,'' \emph{Expert systems with applications}, vol. 186, p. 115736, 2021.

\bibitem{lundberg2017unified}
S.~M. Lundberg and S.-I. Lee, ``A unified approach to interpreting model predictions,'' \emph{Advances in neural information processing systems}, vol.~30, 2017.

\bibitem{84}
A.~Holzinger, A.~Saranti, C.~Molnar, P.~Biecek, and W.~Samek, ``Explainable ai methods-a brief overview,'' in \emph{xxAI-Beyond Explainable AI: International Workshop, Held in Conjunction with ICML 2020, July 18, 2020, Vienna, Austria, Revised and Extended Papers}.\hskip 1em plus 0.5em minus 0.4em\relax Springer, 2022, pp. 13--38.

\bibitem{brahimi2018deep}
M.~Brahimi, M.~Arsenovic, S.~Laraba, S.~Sladojevic, K.~Boukhalfa, and A.~Moussaoui, ``Deep learning for plant diseases: detection and saliency map visualisation,'' \emph{Human and machine learning: Visible, explainable, trustworthy and transparent}, pp. 93--117, 2018.

\bibitem{goldstein2015peeking}
A.~Goldstein, A.~Kapelner, J.~Bleich, and E.~Pitkin, ``Peeking inside the black box: Visualizing statistical learning with plots of individual conditional expectation,'' \emph{journal of Computational and Graphical Statistics}, vol.~24, no.~1, pp. 44--65, 2015.

\bibitem{montavon2019layer}
G.~Montavon, A.~Binder, S.~Lapuschkin, W.~Samek, and K.-R. M{\"u}ller, ``Layer-wise relevance propagation: an overview,'' \emph{Explainable AI: interpreting, explaining and visualizing deep learning}, pp. 193--209, 2019.

\bibitem{quinlan1986induction}
J.~R. Quinlan, ``Induction of decision trees,'' \emph{Machine learning}, vol.~1, pp. 81--106, 1986.

\bibitem{yosinski2015understanding}
J.~Yosinski, J.~Clune, A.~Nguyen, T.~Fuchs, and H.~Lipson, ``Understanding neural networks through deep visualization,'' \emph{arXiv preprint arXiv:1506.06579}, 2015.

\bibitem{fix1985discriminatory}
E.~Fix, \emph{Discriminatory analysis: nonparametric discrimination, consistency properties}.\hskip 1em plus 0.5em minus 0.4em\relax USAF school of Aviation Medicine, 1985, vol.~1.

\bibitem{ribeiro2016should}
M.~T. Ribeiro, S.~Singh, and C.~Guestrin, ``" why should i trust you?" explaining the predictions of any classifier,'' in \emph{Proceedings of the 22nd ACM SIGKDD international conference on knowledge discovery and data mining}, 2016, pp. 1135--1144.

\bibitem{simonyan2013deep}
K.~Simonyan, A.~Vedaldi, and A.~Zisserman, ``Deep inside convolutional networks: Visualising image classification models and saliency maps,'' \emph{arXiv preprint arXiv:1312.6034}, 2013.

\bibitem{bach2015pixel}
S.~Bach, A.~Binder, G.~Montavon, F.~Klauschen, K.-R. M{\"u}ller, and W.~Samek, ``On pixel-wise explanations for non-linear classifier decisions by layer-wise relevance propagation,'' \emph{PloS one}, vol.~10, no.~7, p. e0130140, 2015.

\bibitem{75}
F.~Daudt, D.~Cinalli, and A.~C.~B. Garcia, ``Research on explainable artificial intelligence techniques: An user perspective,'' in \emph{2021 IEEE 24th International Conference on Computer Supported Cooperative Work in Design (CSCWD)}.\hskip 1em plus 0.5em minus 0.4em\relax IEEE, 2021, pp. 144--149.

\bibitem{8}
M.~Ribera and A.~Lapedriza, ``Can we do better explanations? a proposal of user-centered explainable ai.'' in \emph{IUI Workshops}, vol. 2327, 2019, p.~38.

\bibitem{76}
R.~Larasati, ``Explainable ai for breast cancer diagnosis: Application and user’s understandability perception,'' in \emph{2022 International Conference on Electrical, Computer and Energy Technologies (ICECET)}.\hskip 1em plus 0.5em minus 0.4em\relax IEEE, 2022, pp. 1--6.

\bibitem{9}
S.~Dey, P.~Chakraborty, B.~C. Kwon, A.~Dhurandhar, M.~Ghalwash, F.~J.~S. Saiz, K.~Ng, D.~Sow, K.~R. Varshney, and P.~Meyer, ``Human-centered explainability for life sciences, healthcare, and medical informatics,'' \emph{Patterns}, vol.~3, no.~5, p. 100493, 2022.

\bibitem{kim2024stakeholder}
M.~Kim, S.~Kim, J.~Kim, T.-J. Song, and Y.~Kim, ``Do stakeholder needs differ?-designing stakeholder-tailored explainable artificial intelligence (xai) interfaces,'' \emph{International Journal of Human-Computer Studies}, vol. 181, p. 103160, 2024.

\bibitem{77}
M.~K. Lee and K.~Rich, ``Who is included in human perceptions of ai?: Trust and perceived fairness around healthcare ai and cultural mistrust,'' in \emph{Proceedings of the 2021 CHI conference on human factors in computing systems}, 2021, pp. 1--14.

\bibitem{82}
D.~Wang, Q.~Yang, A.~Abdul, and B.~Y. Lim, ``Designing theory-driven user-centric explainable ai,'' in \emph{Proceedings of the 2019 CHI conference on human factors in computing systems}, 2019, pp. 1--15.

\bibitem{81}
Y.~Xie, G.~Gao, and X.~Chen, ``Outlining the design space of explainable intelligent systems for medical diagnosis,'' \emph{arXiv preprint arXiv:1902.06019}, 2019.

\bibitem{80}
C.~J. Cai, S.~Winter, D.~Steiner, L.~Wilcox, and M.~Terry, ``" hello ai": uncovering the onboarding needs of medical practitioners for human-ai collaborative decision-making,'' \emph{Proceedings of the ACM on Human-computer Interaction}, vol.~3, no. CSCW, pp. 1--24, 2019.

\bibitem{78}
M.~Eiband, H.~Schneider, M.~Bilandzic, J.~Fazekas-Con, M.~Haug, and H.~Hussmann, ``Bringing transparency design into practice,'' in \emph{23rd international conference on intelligent user interfaces}, 2018, pp. 211--223.

\bibitem{88}
K.~Niranjan, S.~S. Kumar, S.~Vedanth, and S.~Chitrakala, ``An explainable ai driven decision support system for covid-19 diagnosis using fused classification and segmentation,'' \emph{Procedia Computer Science}, vol. 218, pp. 1915--1925, 2023.

\bibitem{86}
F.~H. Yagin, {\.I}.~B. Cicek, A.~Alkhateeb, B.~Yagin, C.~Colak, M.~Azzeh, and S.~Akbulut, ``Explainable artificial intelligence model for identifying covid-19 gene biomarkers,'' \emph{Computers in Biology and Medicine}, vol. 154, p. 106619, 2023.

\bibitem{basahel2023quantum}
A.~M. Basahel and M.~Yamin, ``Quantum inspired differential evolution with explainable artificial intelligence-based covid-19 detection,'' \emph{Computer Systems Science and Engineering}, pp. 209--224, 2023.

\bibitem{seethi2024explainable}
V.~D.~R. Seethi, Z.~LaCasse, P.~Chivte, J.~Bland, S.~S. Kadkol, E.~R. Gaillard, P.~Bharti, and H.~Alhoori, ``An explainable ai approach for diagnosis of covid-19 using maldi-tof mass spectrometry,'' \emph{Expert Systems with Applications}, vol. 236, p. 121226, 2024.

\bibitem{chadaga2023decision}
K.~Chadaga, S.~Prabhu, V.~Bhat, N.~Sampathila, S.~Umakanth, and R.~Chadaga, ``A decision support system for diagnosis of covid-19 from non-covid-19 influenza-like illness using explainable artificial intelligence,'' \emph{Bioengineering}, vol.~10, no.~4, p. 439, 2023.

\bibitem{kirbouga2023cvd22}
K.~K. K{\i}rbo{\u{g}}a, E.~U. K{\"u}{\c{c}}{\"u}ksille, M.~E. Naldan, M.~I{\c{s}}{\i}k, O.~G{\"u}lc{\"u}, and E.~Aksakal, ``Cvd22: Explainable artificial intelligence determination of the relationship of troponin to d-dimer, mortality, and ck-mb in covid-19 patients,'' \emph{Computer Methods and Programs in Biomedicine}, vol. 233, p. 107492, 2023.

\bibitem{elmannai2023polycystic}
H.~Elmannai, N.~El-Rashidy, I.~Mashal, M.~A. Alohali, S.~Farag, S.~El-Sappagh, and H.~Saleh, ``Polycystic ovary syndrome detection machine learning model based on optimized feature selection and explainable artificial intelligence,'' \emph{Diagnostics}, vol.~13, no.~8, p. 1506, 2023.

\bibitem{khanna2023distinctive}
V.~V. Khanna, K.~Chadaga, N.~Sampathila, S.~Prabhu, V.~Bhandage, and G.~K. Hegde, ``A distinctive explainable machine learning framework for detection of polycystic ovary syndrome,'' \emph{Applied System Innovation}, vol.~6, no.~2, p.~32, 2023.

\bibitem{87}
M.~Hehr, A.~Sadafi, C.~Matek, P.~Lienemann, C.~Pohlkamp, T.~Haferlach, K.~Spiekermann, and C.~Marr, ``Explainable ai identifies diagnostic cells of genetic aml subtypes,'' \emph{PLOS Digital Health}, vol.~2, no.~3, p. e0000187, 2023.

\bibitem{91}
A.~Aghaei and M.~E. Moghaddam, ``Smart roi detection for alzheimer's disease prediction using explainable ai,'' \emph{arXiv preprint arXiv:2303.10401}, 2023.

\bibitem{90}
M.~Zhang, J.~Pan, J.~Lin, M.~Xu, L.~Zhang, R.~Shang, L.~Yao, Y.~Li, W.~Zhou, Y.~Deng \emph{et~al.}, ``An explainable artificial intelligence system for diagnosing helicobacter pylori infection under endoscopy: a case--control study,'' \emph{Therapeutic Advances in Gastroenterology}, vol.~16, p. 17562848231155023, 2023.

\bibitem{tasin2023diabetes}
I.~Tasin, T.~U. Nabil, S.~Islam, and R.~Khan, ``Diabetes prediction using machine learning and explainable ai techniques,'' \emph{Healthcare Technology Letters}, vol.~10, no. 1-2, pp. 1--10, 2023.

\bibitem{albahri2023explainable}
A.~Albahri, S.~S. Joudar, R.~A. Hamid, I.~A. Zahid, M.~Alqaysi, O.~Albahri, A.~Alamoodi, G.~Kou, and I.~M. Sharaf, ``Explainable artificial intelligence multimodal of autism triage levels using fuzzy approach-based multi-criteria decision-making and lime,'' \emph{International Journal of Fuzzy Systems}, pp. 1--30, 2023.

\bibitem{das2023xai}
S.~Das, M.~Sultana, S.~Bhattacharya, D.~Sengupta, and D.~De, ``Xai--reduct: accuracy preservation despite dimensionality reduction for heart disease classification using explainable ai,'' \emph{The Journal of Supercomputing}, pp. 1--31, 2023.

\bibitem{dong2023explainable}
Z.~Dong, J.~Wang, Y.~Li, Y.~Deng, W.~Zhou, X.~Zeng, D.~Gong, J.~Liu, J.~Pan, R.~Shang \emph{et~al.}, ``Explainable artificial intelligence incorporated with domain knowledge diagnosing early gastric neoplasms under white light endoscopy,'' \emph{NPJ Digital Medicine}, vol.~6, no.~1, p.~64, 2023.

\bibitem{auzine2023classification}
M.~M. Auzine, M.~H.-M. Khan, S.~Baichoo, N.~G. Sahib, X.~Gao, and P.~Bissoonauth-Daiboo, ``Classification of gastrointestinal cancer through explainable ai and ensemble learning,'' in \emph{2023 Sixth International Conference of Women in Data Science at Prince Sultan University (WiDS PSU)}.\hskip 1em plus 0.5em minus 0.4em\relax IEEE, 2023, pp. 195--200.

\bibitem{volkov2023possibilities}
E.~N. Volkov and A.~N. Averkin, ``Possibilities of explainable artificial intelligence for glaucoma detection using the lime method as an example,'' in \emph{2023 XXVI International Conference on Soft Computing and Measurements (SCM)}.\hskip 1em plus 0.5em minus 0.4em\relax IEEE, 2023, pp. 130--133.

\bibitem{adilakshmi2023medical}
J.~Adilakshmi, G.~V. Reddy, K.~D. Nidumolu, R.~D.~C. Pecho, and M.~J. Pasha, ``A medical diagnosis system based on explainable artificial intelligence: Autism spectrum disorder diagnosis,'' \emph{International Journal of Intelligent Systems and Applications in Engineering}, vol.~11, no.~6s, pp. 385--402, 2023.

\bibitem{ahmed2023identification}
F.~Ahmed, M.~Asif, M.~Saleem, U.~F. Mushtaq, and M.~Imran, ``Identification and prediction of brain tumor using vgg-16 empowered with explainable artificial intelligence,'' \emph{International Journal of Computational and Innovative Sciences}, vol.~2, no.~2, pp. 24--33, 2023.

\bibitem{ramirez2023explainable}
A.~Ram{\'\i}rez-Mena, E.~Andr{\'e}s-Le{\'o}n, M.~J. Alvarez-Cubero, A.~Anguita-Ruiz, L.~J. Martinez-Gonzalez, and J.~Alcala-Fdez, ``Explainable artificial intelligence to predict and identify prostate cancer tissue by gene expression,'' \emph{Computer Methods and Programs in Biomedicine}, vol. 240, p. 107719, 2023.

\bibitem{panesar2019artificial}
S.~Panesar, Y.~Cagle, D.~Chander, J.~Morey, J.~Fernandez-Miranda, and M.~Kliot, ``Artificial intelligence and the future of surgical robotics,'' \emph{Annals of surgery}, vol. 270, no.~2, pp. 223--226, 2019.

\bibitem{england2019artificial}
J.~R. England and P.~M. Cheng, ``Artificial intelligence for medical image analysis: a guide for authors and reviewers,'' \emph{American journal of roentgenology}, vol. 212, no.~3, pp. 513--519, 2019.

\bibitem{reverberi2022experimental}
C.~Reverberi, T.~Rigon, A.~Solari, C.~Hassan, P.~Cherubini, and A.~Cherubini, ``Experimental evidence of effective human--ai collaboration in medical decision-making,'' \emph{Scientific reports}, vol.~12, no.~1, p. 14952, 2022.

\bibitem{croatti2019bdi}
A.~Croatti, S.~Montagna, A.~Ricci, E.~Gamberini, V.~Albarello, and V.~Agnoletti, ``Bdi personal medical assistant agents: The case of trauma tracking and alerting,'' \emph{Artificial intelligence in medicine}, vol.~96, pp. 187--197, 2019.

\bibitem{26}
A.~Holzinger, C.~Biemann, C.~S. Pattichis, and D.~B. Kell, ``What do we need to build explainable ai systems for the medical domain?'' \emph{arXiv preprint arXiv:1712.09923}, 2017.

\bibitem{27}
A.~Lucieri, M.~N. Bajwa, A.~Dengel, and S.~Ahmed, ``Achievements and challenges in explaining deep learning based computer-aided diagnosis systems,'' \emph{arXiv preprint arXiv:2011.13169}, 2020.

\bibitem{hoerbst2010electronic}
A.~Hoerbst and E.~Ammenwerth, ``Electronic health records,'' \emph{Methods of information in medicine}, vol.~49, no.~04, pp. 320--336, 2010.

\bibitem{khedkar2019explainable}
S.~Khedkar, V.~Subramanian, G.~Shinde, and P.~Gandhi, ``Explainable ai in healthcare,'' in \emph{Healthcare (April 8, 2019). 2nd International Conference on Advances in Science \& Technology (ICAST)}, 2019.

\bibitem{duell2021comparison}
J.~Duell, X.~Fan, B.~Burnett, G.~Aarts, and S.-M. Zhou, ``A comparison of explanations given by explainable artificial intelligence methods on analysing electronic health records,'' in \emph{2021 IEEE EMBS International Conference on Biomedical and Health Informatics (BHI)}.\hskip 1em plus 0.5em minus 0.4em\relax IEEE, 2021, pp. 1--4.

\bibitem{lauritsen2020explainable}
S.~M. Lauritsen, M.~Kristensen, M.~V. Olsen, M.~S. Larsen, K.~M. Lauritsen, M.~J. J{\o}rgensen, J.~Lange, and B.~Thiesson, ``Explainable artificial intelligence model to predict acute critical illness from electronic health records,'' \emph{Nature communications}, vol.~11, no.~1, p. 3852, 2020.

\bibitem{fan2021interpretability}
F.-L. Fan, J.~Xiong, M.~Li, and G.~Wang, ``On interpretability of artificial neural networks: A survey,'' \emph{IEEE Transactions on Radiation and Plasma Medical Sciences}, vol.~5, no.~6, pp. 741--760, 2021.

\bibitem{100}
P.~R. Magesh, R.~D. Myloth, and R.~J. Tom, ``An explainable machine learning model for early detection of parkinson's disease using lime on datscan imagery,'' \emph{Computers in Biology and Medicine}, vol. 126, p. 104041, 2020.

\bibitem{hinton2015distilling}
G.~Hinton, O.~Vinyals, and J.~Dean, ``Distilling the knowledge in a neural network,'' \emph{arXiv preprint arXiv:1503.02531}, 2015.

\bibitem{93}
N.~Prentzas, A.~Nicolaides, E.~Kyriacou, A.~Kakas, and C.~Pattichis, ``Integrating machine learning with symbolic reasoning to build an explainable ai model for stroke prediction,'' in \emph{2019 IEEE 19th International Conference on Bioinformatics and Bioengineering (BIBE)}.\hskip 1em plus 0.5em minus 0.4em\relax IEEE, 2019, pp. 817--821.

\bibitem{chaddad2023texture}
A.~Chaddad, L.~Hassan, and Y.~Katib, ``A texture-based method for predicting molecular markers and survival outcome in lower grade glioma,'' \emph{Applied Intelligence}, pp. 1--15, 2023.

\bibitem{feng2021can}
L.~Feng, S.~Shu, Z.~Lin, F.~Lv, L.~Li, and B.~An, ``Can cross entropy loss be robust to label noise?'' in \emph{Proceedings of the twenty-ninth international conference on international joint conferences on artificial intelligence}, 2021, pp. 2206--2212.

\bibitem{kingma2014adam}
D.~P. Kingma and J.~Ba, ``Adam: A method for stochastic optimization,'' \emph{arXiv preprint arXiv:1412.6980}, 2014.

\bibitem{msoud_nickparvar_2021}
M.~Nickparvar, ``Chest ct-scan images dataset,'' \url{https://www.kaggle.com/datasets/masoudnickparvar/brain-tumor-mri-dataset}, 2021.

\bibitem{MOHAMED_HANY}
M.~Hany, ``Chest ct-scan images dataset,'' \url{https://www.kaggle.com/datasets/mohamedhanyyy/chest-ctscan-images}, 2020.

\bibitem{FRANCIS_JESMAR_MONTALBO}
F.~J. Montalbo, ``Wce curated colon disease dataset,'' \url{https://www.kaggle.com/datasets/francismon/curated-colon-dataset-for-deep-learning}, 2022.

\bibitem{GUNA_VENKAT_DODDI_2022}
G.~V. Doddi, ``eye diseases classification,'' \url{https://www.kaggle.com/datasets/gunavenkatdoddi/eye-diseases-classification}, 2022.

\bibitem{PRASHANT_PATEL_2021}
P.~Patel, ``Chest x-ray (covid-19 \& pneumonia),'' \url{https://www.kaggle.com/datasets/prashant268/chest-xray-covid19-pneumonia}, 2021.

\bibitem{dwivedi2024efficient}
R.~Dwivedi, P.~Kothari, D.~Chopra, M.~Singh, and R.~Kumar, ``An efficient ensemble explainable ai (xai) approach for morphed face detection,'' \emph{Pattern Recognition Letters}, vol. 184, pp. 197--204, 2024.

\bibitem{nauta2023anecdotal}
M.~Nauta, J.~Trienes, S.~Pathak, E.~Nguyen, M.~Peters, Y.~Schmitt, J.~Schl{\"o}tterer, M.~van Keulen, and C.~Seifert, ``From anecdotal evidence to quantitative evaluation methods: A systematic review on evaluating explainable ai,'' \emph{ACM Computing Surveys}, vol.~55, no. 13s, pp. 1--42, 2023.

\bibitem{haque2023explainable}
A.~B. Haque, A.~N. Islam, and P.~Mikalef, ``Explainable artificial intelligence (xai) from a user perspective: A synthesis of prior literature and problematizing avenues for future research,'' \emph{Technological Forecasting and Social Change}, vol. 186, p. 122120, 2023.

\bibitem{alizadehsani2024explainable}
R.~Alizadehsani, S.~S. Oyelere, S.~Hussain, S.~K. Jagatheesaperumal, R.~R. Calixto, M.~Rahouti, M.~Roshanzamir, and V.~H.~C. De~Albuquerque, ``Explainable artificial intelligence for drug discovery and development-a comprehensive survey,'' \emph{IEEE Access}, 2024.

\bibitem{lipton2018mythos}
Z.~C. Lipton, ``The mythos of model interpretability: In machine learning, the concept of interpretability is both important and slippery.'' \emph{Queue}, vol.~16, no.~3, pp. 31--57, 2018.

\bibitem{105}
A.~Albahri, A.~M. Duhaim, M.~A. Fadhel, A.~Alnoor, N.~S. Baqer, L.~Alzubaidi, O.~Albahri, A.~Alamoodi, J.~Bai, A.~Salhi \emph{et~al.}, ``A systematic review of trustworthy and explainable artificial intelligence in healthcare: Assessment of quality, bias risk, and data fusion,'' \emph{Information Fusion}, 2023.

\bibitem{106}
D.~W. Joyce, A.~Kormilitzin, K.~A. Smith, and A.~Cipriani, ``Explainable artificial intelligence for mental health through transparency and interpretability for understandability,'' \emph{npj Digital Medicine}, vol.~6, no.~1, p.~6, 2023.

\bibitem{107}
O.~Wysocki, J.~K. Davies, M.~Vigo, A.~C. Armstrong, D.~Landers, R.~Lee, and A.~Freitas, ``Assessing the communication gap between ai models and healthcare professionals: explainability, utility and trust in ai-driven clinical decision-making,'' \emph{Artificial Intelligence}, vol. 316, p. 103839, 2023.

\bibitem{108}
T.~Dhar, N.~Dey, S.~Borra, and R.~S. Sherratt, ``Challenges of deep learning in medical image analysis-improving explainability and trust,'' \emph{IEEE Transactions on Technology and Society}, 2023.

\bibitem{109}
W.~Jin, X.~Li, M.~Fatehi, and G.~Hamarneh, ``Guidelines and evaluation of clinical explainable ai in medical image analysis,'' \emph{Medical Image Analysis}, vol.~84, p. 102684, 2023.

\bibitem{viswan2024explainable}
V.~Viswan, N.~Shaffi, M.~Mahmud, K.~Subramanian, and F.~Hajamohideen, ``Explainable artificial intelligence in alzheimer’s disease classification: A systematic review,'' \emph{Cognitive Computation}, vol.~16, no.~1, pp. 1--44, 2024.

\bibitem{mohseni2021multidisciplinary}
S.~Mohseni, N.~Zarei, and E.~D. Ragan, ``A multidisciplinary survey and framework for design and evaluation of explainable ai systems,'' \emph{ACM Transactions on Interactive Intelligent Systems (TiiS)}, vol.~11, no. 3-4, pp. 1--45, 2021.

\bibitem{ueda2024fairness}
D.~Ueda, T.~Kakinuma, S.~Fujita, K.~Kamagata, Y.~Fushimi, R.~Ito, Y.~Matsui, T.~Nozaki, T.~Nakaura, N.~Fujima \emph{et~al.}, ``Fairness of artificial intelligence in healthcare: review and recommendations,'' \emph{Japanese Journal of Radiology}, vol.~42, no.~1, pp. 3--15, 2024.

\bibitem{vilone2021notions}
G.~Vilone and L.~Longo, ``Notions of explainability and evaluation approaches for explainable artificial intelligence,'' \emph{Information Fusion}, vol.~76, pp. 89--106, 2021.

\bibitem{patra2024xai}
G.~Patra and S.~Datta, ``Xai for society 5.0: Requirements, opportunities, and challenges in the current context,'' \emph{XAI Based Intelligent Systems for Society 5.0}, pp. 269--293, 2024.

\bibitem{34}
L.~Vigano and D.~Magazzeni, ``Explainable security,'' in \emph{2020 IEEE European Symposium on Security and Privacy Workshops (EuroS\&PW)}.\hskip 1em plus 0.5em minus 0.4em\relax IEEE, 2020, pp. 293--300.

\bibitem{35}
T.~Miller, ``Explanation in artificial intelligence: Insights from the social sciences,'' \emph{Artificial intelligence}, vol. 267, pp. 1--38, 2019.

\bibitem{36}
F.~Yang, M.~Du, and X.~Hu, ``Evaluating explanation without ground truth in interpretable machine learning,'' \emph{arXiv preprint arXiv:1907.06831}, 2019.

\end{thebibliography}
}

\end{document}